\documentclass[sigconf]{acmart}

\AtBeginDocument{%
  \providecommand\BibTeX{{%
    \normalfont B\kern-0.5em{\scshape i\kern-0.25em b}\kern-0.8em\TeX}}}

\copyrightyear{2021}
\acmYear{2021}
\setcopyright{acmcopyright}\acmConference[KDD '21]{Proceedings of the 27th ACM SIGKDD Conference on Knowledge Discovery and Data Mining}{August 14--18, 2021}{Virtual Event, Singapore}
\acmBooktitle{Proceedings of the 27th ACM SIGKDD Conference on Knowledge Discovery and Data Mining (KDD '21), August 14--18, 2021, Virtual Event, Singapore} \acmPrice{15.00}
\acmDOI{10.1145/3447548.3467061}
\acmISBN{978-1-4503-8332-5/21/08}

\usepackage{subfigure}
\usepackage{multirow}
\usepackage{bm}
\usepackage{bbm}
\usepackage{dsfont}
\usepackage{color,soul}
\usepackage{listings}
\usepackage{xcolor}

\usepackage[linesnumbered,vlined,ruled,noend]{algorithm2e}
\usepackage{xspace}
\newcommand{\sys}{\textsc{OpenBox}\xspace}

\newcommand{\red}[1]{\textcolor{red}{#1}}
\newcommand{\blue}[1]{\textcolor{blue}{#1}}

\newcommand{\para}[1]{{\vspace{2pt} \bf \noindent #1 \hspace{1pt}}}
\definecolor{codegray}{rgb}{0.5,0.5,0.5}

\usepackage{enumitem}

\begin{document}
\fancyhead{}
\title{\sys: A Generalized Black-box Optimization Service}

\author{Yang Li$^{\dagger}$, Yu Shen$^{\dagger\mathsection}$, Wentao Zhang$^\dagger$,
Yuanwei Chen$^\dagger$, 
Huaijun Jiang$^{\dagger\mathsection}$, 
Mingchao Liu$^\dagger$}
\author{
Jiawei Jiang$^\ddagger$,
Jinyang Gao$^\diamond$, Wentao Wu$^*$, Zhi Yang$^\dagger$, Ce Zhang$^\ddagger$, Bin Cui$^{\dagger\nabla}$}

\affiliation{
$^\dagger$
Key Laboratory of High Confidence Software Technologies (MOE), School of EECS, Peking University\country{China}
}

\affiliation{
$^\ddagger$Department of Computer Science, Systems Group, ETH Zurich\country{Switzerland}
}

\affiliation{
$^\nabla$Institute of Computational Social Science, Peking University (Qingdao)\country{China}
}

\affiliation{
$^*$Microsoft Research\country{USA}~~~~~
$^\diamond$Alibaba Group\country{China}
~~~~~$^\mathsection$Kuaishou Technology\country{China}
}

\affiliation{
$^\dagger$\{liyang.cs, shenyu, wentao.zhang,  yw.chen,  jianghuaijun, by\_liumingchao, yangzhi, bin.cui\}@pku.edu.cn 
\\
$^\ddagger$\{jiawei.jiang, ce.zhang\}@inf.ethz.ch~~~~~$^*$wentao.wu@microsoft.com~~~~~$^\Diamond$jinyang.gjy@alibaba-inc.com\country{}
}\country{}

\renewcommand{\shortauthors}{Li et al.}

\begin{abstract}
Black-box optimization (BBO) has a broad range of applications, including automatic machine learning, engineering, physics, and experimental design.
However, it remains a challenge for users to apply BBO methods to their problems at hand with existing software packages, in terms of applicability, performance, and efficiency. 
In this paper, we build \sys, an open-source and general-purpose BBO service with \emph{improved usability}.
The modular design behind \sys also facilitates flexible abstraction and optimization of basic BBO components that are common in other existing systems.
\sys is distributed, fault-tolerant, and scalable.
To improve efficiency, \sys further utilizes ``algorithm agnostic'' parallelization and transfer learning.
Our experimental results demonstrate the effectiveness and efficiency of \sys compared to existing systems.

\end{abstract}





\settopmatter{printfolios=true}
\maketitle

\section{Introduction}

Black–box optimization (BBO) is the task of optimizing an objective function within a limited budget for function evaluations.
``Black-box'' means that the objective function has no analytical form so that information such as the derivative of the objective function is unavailable.
Since the evaluation of objective functions is often expensive, the goal of black-box optimization is to find a configuration that approaches the global optimum as rapidly as possible.

\begin{table}[t]
    \scriptsize
    \centering
    \begin{tabular}{l|ccccc}
        \hline
        System/Package & Multi-obj. & FIOC & Constraint & History & Distributed \\
        \hline
        Hyperopt & $\times$ & \checkmark & $\times$ & $\times$ &  \checkmark \\
        Spearmint & $\times$ & $\times$ & \checkmark & $\times$ &  $\times$ \\
        SMAC3 & $\times$ & \checkmark & $\times$ & $\times$ & $\times$ \\
        BoTorch & \checkmark & $\times$ & \checkmark & $\times$ & $\times$ \\
        GPflowOpt & \checkmark & $\times$ & \checkmark & $\times$ & $\times$ \\
        Vizier & $\times$ & \checkmark & $\times$ & $\triangle$ & \checkmark \\
        HyperMapper & \checkmark & \checkmark & \checkmark & $\times$ & $\times$ \\
        HpBandSter & $\times$ & \checkmark & $\times$ & $\times$ & \checkmark \\
        \hline
        \textbf{\sys} & \checkmark & \checkmark & \checkmark & \checkmark & \checkmark\\ 
        \hline
    \end{tabular}
    \vspace{1em}
    \caption{A taxonomy of BBO systems/softwares.
    \textit{Multi-obj.} notes whether the system supports multiple objectives or not. 
    \textit{FIOC} indicates if the system supports all Float, Integer, Ordinal and
    Categorical variables. 
    \textit{Constraint} refers to the support for inequality constraints. 
    \textit{History} represents the ability of the system to inject the prior knowledge from previous tasks in the search.
    \textit{Distributed} notes if it supports parallel evaluations under a distributed environment.
    $\triangle$ means the system cannot support it for many cases. Note that, \textsc{BoTorch}, as a framework, might provide the algorithmic building blocks 
    for a developer to \textit{implement} some of these capacities.}
    \label{tbl:sys_cmp}
\end{table}

Traditional BBO with a single objective has many applications: 
1) automatic A/B testing, 2) experimental design~\cite{foster2019variational}, 3) knobs tuning in database~\cite{van2017automatic,zhang2019end,Zhang2021FacilitatingDT}, and 4) automatic hyper-parameter tuning~\cite{hutter2011sequential,snoek2012practical,bergstra2011algorithms,li2020efficient}, one of the most indispensable components in AutoML systems~\cite{20.500.11850/458916,Li2021VolcanoMLSU} such as
Microsoft's Azure Machine Learning, Google's Cloud Machine Learning, Amazon Machine Learning~\cite{liberty2020elastic}, and IBM’s Watson Studio AutoAI, where the task is to minimize the validation error of a machine learning algorithm as a function of its hyper-parameters.
Recently, \emph{generalized} BBO emerges and has been applied to many areas such as 1) processor architecture and circuit design~\cite{Azizi2010}, 2) resource allocation~\cite{Gelbart2015}, and 3) automatic chemical design~\cite{Ryan2020}, which requires more general functionalities that may not be supported by traditional BBO, such as multiple objectives and constraints.
As examples of applications of generalized BBO in the software industry, Microsoft's Smart Buildings project~\cite{ms_smartbuilding} searches for the best smart building designs by minimizing both energy consumption and construction costs (i.e., BBO with multiple objectives); Amazon Web Service aims to optimize the performance of machine learning models while enforcing fairness constraints~\cite{perrone2020fair} (i.e., BBO with constraints).



\begin{figure}[t]
\centering
\scalebox{0.6}[0.6]{
\includegraphics[width=0.5\textwidth]{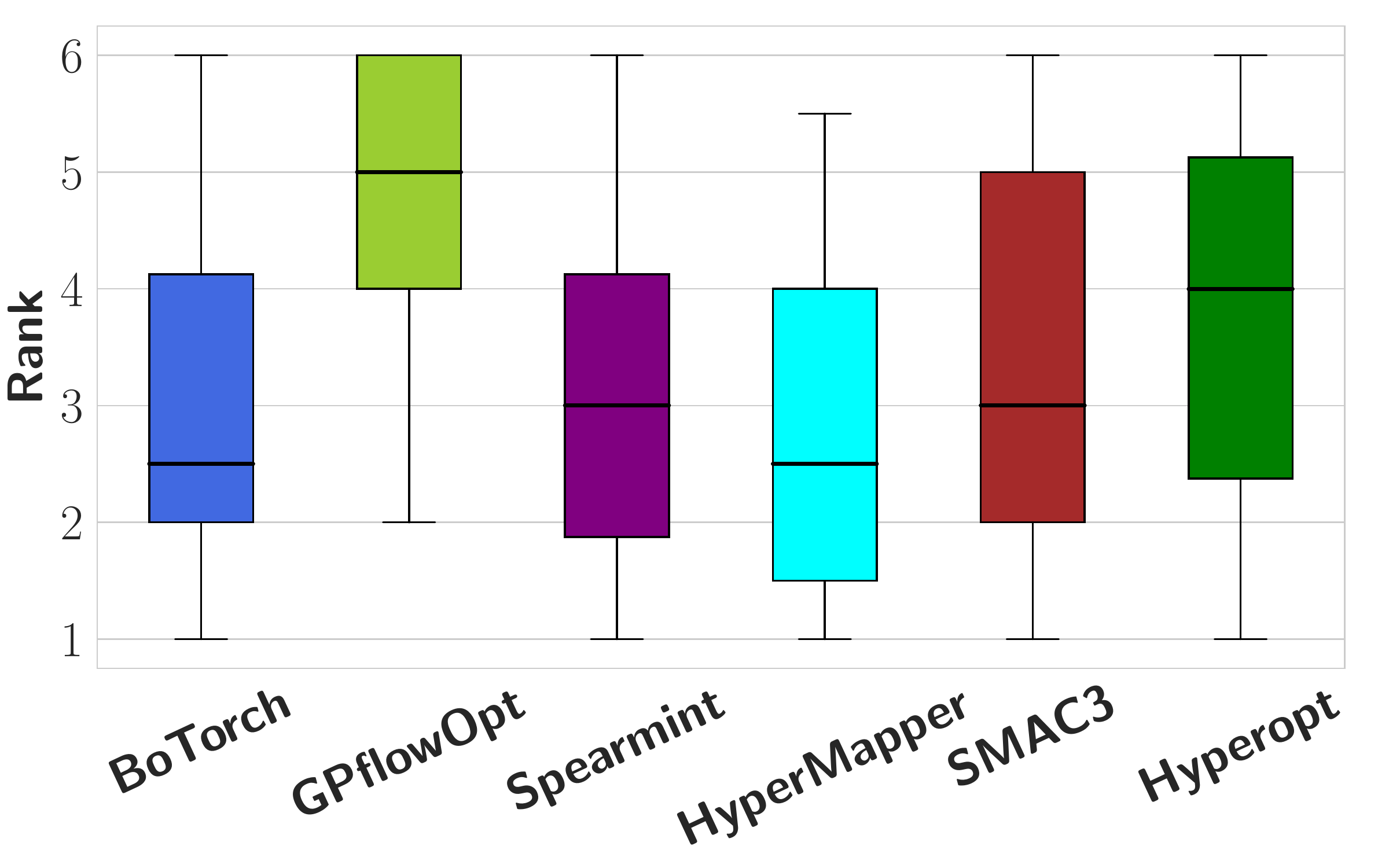}
}
\caption{Performance rank of softwares on 25 AutoML tasks (\textit{lower is better}). The box extends from the lower to the upper quartile values, with a line at the median. The whiskers that extend the box show the range of the data.}
\label{fig:software_cmp}
\end{figure}

Many software packages and platforms have been developed for traditional BBO (see Table~\ref{tbl:sys_cmp}).
Yet, to the best of our knowledge, so far there is no platform that is designed to target generalized BBO.
The existing BBO packages have the following three limitations when applied to general BBO scenarios:


\noindent(1) \underline{Restricted scope and applicability.} Restricted by the underlying algorithms, most existing BBO implementations cannot handle diverse optimization problems in a unified manner (see Table~\ref{tbl:sys_cmp}). 
For example, \texttt{Hyperopt}~\cite{bergstra2011algorithms}, \texttt{SMAC3}~\cite{hutter2011sequential}, and \texttt{HpBandSter}~\cite{falkner2018bohb} can only deal with single-objective problems without constraints.
Though \texttt{BoTorch}~\cite{balandat2020botorch} and \texttt{GPflowOpt}~\cite{GPflowOpt2017} 
can be used, as a framework, for developers to implement new optimization problems with multi-objectives and constraints; nevertheless, their current off-the-shelf 
supports are also limited (e.g., the support for non-continuous parameters).
\newline(2) \underline{Unstable performance across problems.}
Most existing software packages only implement one or very few BBO algorithms. 
According to the ``no free lunch'' theorem~\cite{ho2001simple}, no single algorithm can achieve the best performance for all BBO problems.
Therefore, existing packages would inevitably suffer from unstable performance when applied to different problems.
Figure~\ref{fig:software_cmp} presents a brief example of hyper-parameter tuning across 25 AutoML tasks, where for each problem we rank the packages according to their performances.
We can observe that all packages exhibit unstable performance, and no one consistently outperforms the others.
This poses challenges on practitioners to select the best package for a specific problem, 
which usually requires deep domain knowledge/expertise and is typically very time-consuming.
\newline(3) \underline{Limited scalability and efficiency.} 
Most existing packages execute optimization in a sequential manner, which is inherently inefficient and unscalable. However, extending the sequential algorithm to make it parallelizable is nontrivial and requires significant engineering efforts.
Moreover, most existing systems cannot support \emph{transfer learning} to accelerate the optimization on a similar task.

With these challenges, in this paper we propose \sys, a system for \emph{generalized} black-box optimization.
The design of \sys follows the philosophy of providing ``BBO as a service'' --- 
instead of developing another software package, we opt to implement \sys as a distributed, fault-tolerant, scalable, and efficient \emph{service}, which addresses the aforementioned challenges in a uniform manner and brings additional advantages such as ease of use, portability, and zero maintenance.
In this regard, Google's \texttt{Vizier}~\cite{golovin2017google} is perhaps the only existing BBO service as far as we know that follows the same design philosophy.
Nevertheless, \texttt{Vizier} only supports traditional BBO, and cannot be applied to general scenarios with multiple objectives and constraints that \sys aims for.
Moreover, unlike \texttt{Vizier}, which remains Google's internal service as of today, we have open-sourced \sys that is available at \url{https://github.com/PKU-DAIR/open-box}.


The key novelty of \sys lies in both the system implementation and algorithm design. 
In terms of system implementation, \sys allows users to define their tasks and access the generalized BBO service conveniently via a task description language (TDL) along with customized interfaces.
\sys also introduces a high-level parallel mechanism by decoupling basic components in common optimization algorithms, which is ``algorithm agnostic'' and enables parallel execution in both synchronous and asynchronous settings.
Moreover, \sys also provides a general transfer-learning framework for generalized BBO, which can leverage the prior knowledge acquired from previous tasks to improve the efficiency of the current optimization task.
In terms of algorithm design, \sys can host most of the state-of-the-art optimization algorithms and make their performances more \emph{stable} via an \emph{automatic} algorithm selection module, which can choose proper optimization algorithm for a given problem automatically.
Furthermore, \sys also supports \emph{multi-fidelity} and \emph{early-stopping} algorithms for further optimization of algorithm efficiency.


\vspace{0.5em}
\noindent
{\bf Contributions.} In summary, our main contributions are:

\noindent
\textit{C1. An open-sourced service for generalized BBO.}
To the best of our knowledge, \sys is the first open-sourced service for efficient and general black-box optimization.

\noindent
\textit{C2. Ease of use.} \sys provides user-friendly interfaces, visualization, resource-aware management, and automatic algorithm selection for consistent performance.

\noindent
\textit{C3. High efficiency and scalability.} 
We develop scalable and general frameworks for transfer-learning and distributed parallel execution in \sys. 
These building blocks are properly integrated to handle diverse optimization scenarios efficiently.

\noindent
\textit{C4. State-of-the-art performance.}
Our empirical evaluation demonstrates that \sys achieves state-of-the-art performance compared to existing systems over a wide range of BBO tasks.

\vspace{0.5em}
\noindent
{\bf Moving Forward.} With the above advantages and features, \sys can be used for optimizing a wide variety of different applications in an industrial setting. We are currently conducting an initial deployment of \sys in \href{https://www.kuaishou.com/en/}{Kuaishou}, one of the most popular ``short video'' platforms in China, to automate the tedious process
of hyperparameter tuning. Initial results have suggested we can outperform human experts.

\section{Background and Related Work}

\para{Generalized Black-box Optimization (BBO).}
Black-box optimization makes few assumptions about the problem, and is thus applicable in a wide range of scenarios. 
We define the generalized BBO problem as follows.
The objective function of generalized BBO is a vector-valued black-box function $\bm{f}(\bm{x}):\mathcal{X} \rightarrow \mathbbm{R}^{p}$, where $\mathcal{X}$ is the search space of interest. 
The goal is to identify the set of \textit{Pareto optimal} solutions $\mathcal{P}^* = \{\bm{f}(\bm{x}) \text{ s.t. } \nexists \ \bm{x'} \in \mathcal{X} : \bm{f}(\bm{x'}) \prec \bm{f}(\bm{x})\}$, such that any improvement in one objective means deteriorating another. To approximate $\mathcal{P}^*$, we compute the finite Pareto set $\mathcal{P}$ from observed data $\{(\bm{x_i}, \bm{y_i})\}_{i=1}^n$.
When $p = 1$, the problem becomes single-objective BBO, as $\mathcal{P} = \{y_{\text{best}}\}$ where $y_{\text{best}}$ is defined as the best objective value observed. 
We also consider the case with black-box inequality constraints. Denote the set of feasible points by $\mathcal{C} = \{ \bm{x} : c_1(\bm{x}) \leq 0, \ldots, c_q(\bm{x}) \leq 0 \}$. Under this setting, we aim to identify the feasible Pareto set $\mathcal{P}_{\text{feas}} = \{\bm{f}(\bm{x}) \text{ s.t. } \bm{x} \in \mathcal{C}, \ \nexists \ \bm{x'} \in \mathcal{X} : \bm{f}(\bm{x'}) \prec \bm{f}(\bm{x}),\ \bm{x'} \in \mathcal{C}\}$.

\para{Black-box Optimization Methods.}
Black-box optimization has been studied extensively in many fields, including derivative-free optimization~\cite{Rios2013},
Bayesian optimization (BO)~\cite{bo_survey}, evolutionaray algorithms~\cite{Hansen2001}, multi-armed bandit algorithms~\cite{Srinivas2010,li2018hyperband}, etc.
To optimize expensive-to-evaluate black-box functions with as few evaluations as possible, \sys adopts BO, one of the most prevailing frameworks in BBO, as the basic optimization framework.
BO iterates between fitting probabilistic surrogate models and determining which configuration to evaluate next by maximizing an acquisition function. With different choices of acquisition functions, BO can be applied to generalized BBO problems.

\textit{BBO with Multiple Objectives.}
Many multi-objective BBO algorithms have been proposed~\cite{Knowles2006, paria2020flexible, belakaria2020uncertainty,  Lobato2016, Belakaria2019}.
Couckuyt et. al.~\cite{Couckuyt2014} propose the Hypervolume Probability of Improvement (HVPOI); Yang et. al.~\cite{yang2019multi} and Daulton et. al.~\cite{daulton2020differentiable} use the Expected Hypervolume Improvement (EHVI) metrics.

\textit{BBO with Black-box Constraints.}
Gardner et.al.~\cite{Gardner2014} present Probability of Feasibility (PoF), which uses GP surrogates to model the constraints. 
In general, multiplying PoF with the unconstrained acquisition function produces the constrained version of it. 
SCBO~\cite{eriksson2021scalable} employs the trust region method and scales to large batches by extending Thompson sampling to constrained optimization. Other methods handle constraints in different ways~\cite{gramacy2016modeling, picheny2016bayesian, hernandez2015predictive}.
For multi-objective optimization with constraints, PESMOC~\cite{garrido2019predictive} and MESMOC~\cite{belakaria2020uncertainty} support constraints by adding the entropy of the conditioned predictive distribution.

\para{BBO Systems and Packages.}
Many of these algorithms have available open-source implementations. \texttt{BoTorch}, \texttt{GPflowOpt} and \texttt{HyperMapper} implement several BO algorithms to solve mathematical problems in different settings. 
Within the machine learning community, \texttt{Hyperopt}, \texttt{Spearmint}, \texttt{SMAC3} and \texttt{HpBandSter} aim to optimize the hyper-parameters of machine learning models. 
Google's \texttt{Vizier} is one of the early attempts in building service for BBO.
We also note that \textsc{Facebook Ax}\footnote{\url{https://github.com/facebook/ax}}  provides high-level API for BBO with \textsc{BoTorch} as its Bayesian optimization engine.

\section{System Overview}

\begin{figure}[t]
\centering
\includegraphics[width=0.48\textwidth]{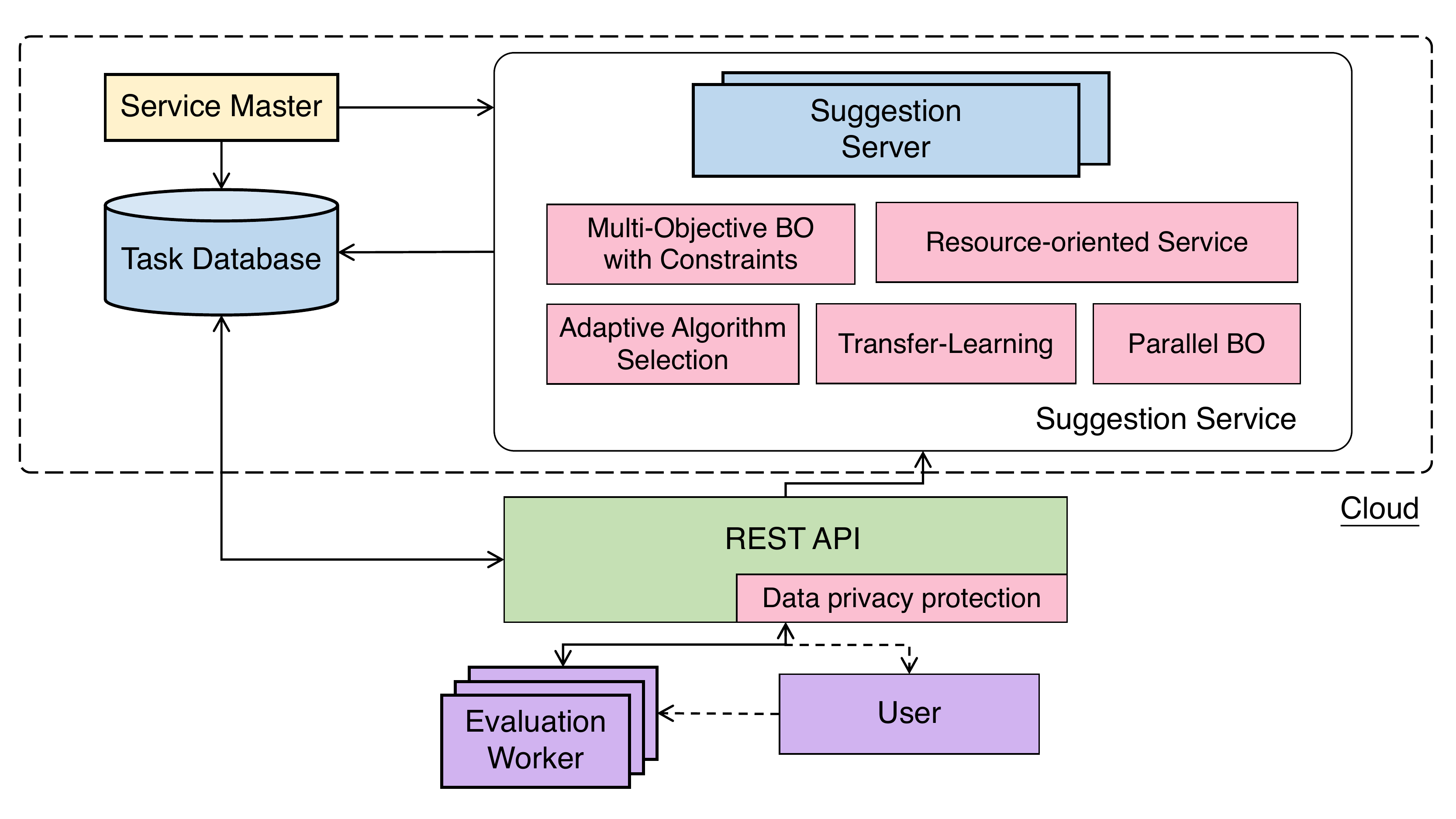}
\vspace{-2.2em}
\caption{Architecture of \sys.}
\label{fig:sys_architecture}
\end{figure}

In this section, we provide the basic concepts in the paper, explore the design principles in implementing black-box optimization (BBO) as a service, and describe the system architecture.

\subsection{Definitions}
Throughout the paper, we use the following terms to describe the semantics of the system:

\noindent
\underline{\emph{Configuration.}} 
Also called suggestion, a vector $\bm{x}$ sampled from the given search space $\mathcal{X}$; each element in $\bm{x}$ is an assignment of a parameter from its domain.

\noindent
\underline{\emph{Trial.}} 
Corresponds to an evaluation of a configuration $\bm{x}$, which has three status: Completed, Running, Ready. Once a trial is completed, we can obtain the evaluation result $\bm{f}(\bm{x})$.

\noindent
\underline{\emph{Task.}} 
A BBO problem over a search space $\mathcal{X}$.
The task type is identified by the number of objectives and constraints.

\noindent
\underline{\emph{Worker.}} 
Refers to a process responsible for executing a trial.

\subsection{Goals and Principles}

\subsubsection{\underline{Design Goal}}

As mentioned before, \sys 's design satisfies the following desiderata:
\begin{itemize}
    \item \textit{Ease of use.} Minimal user effort, and user-friendly visualization for tracking and managing BBO tasks.
    \item \textit{Consistent performance.} Host state-of-the-art optimization algorithms; choose the proper algorithm automatically.
    \item \textit{Resource-aware management.} Give cost-model based advice to users, e.g., minimal workers or time-budget.
    \item \textit{Scalability.} Scale to dimensions on the number of input variables, objectives, tasks, trials, and parallel evaluations.
    \item \textit{High efficiency.} Effective use of parallel resources, system optimization with transfer-learning and multi-fidelities, etc.
    \item \textit{Fault tolerance, extensibility, and data privacy protection.}
\end{itemize}

\subsubsection{\underline{Design Principles}}
We present the key principles underlying the design of \sys.

\textbf{P1: Provide convenient service API that abstracts the implementation and execution complexity away from the user.} For ease of use, we adopt the ``\textit{BBO as a service}'' paradigm and implement \sys as a managed general service for black-box optimization. 
Users can access this service via \texttt{REST API} conveniently (see Figure~\ref{fig:sys_architecture}), and do not need to worry about other issues such as environment setup, software maintenance, programming, and optimization of the execution.
Moreover, we also provide a \texttt{Web UI}, through which users can easily track and manage the tasks.

\textbf{P2: Separate optimization algorithm selection complexity away from the user.}
Users do not need to disturb themselves with choosing the proper algorithm to solve a specific problem via the automatic algorithm selection module.
Furthermore, an important decision is to keep our service \emph{stateless} (see Figure~\ref{fig:sys_architecture}), so that we can seamlessly switch algorithms during a task, i.e., dynamically choose the algorithm that is likely to perform the best for a particular task.
This enables \sys to achieve satisfactory performance once the BBO algorithm is selected properly.

\textbf{P3: Support general distributed parallelization and transfer learning.}
We aim to provide users with full potential to improve the efficiency of the BBO service. 
We design an ``algorithm agnostic'' mechanism that can parallelize the BBO algorithms (Sec. \ref{lp_based_parallel}),
through which we do not need to re-design the parallel version for each algorithm individually. 
Moreover, if the optimization history over similar tasks is provided, our transfer learning framework can leverage the history to accelerate the current task (Sec. \ref{tl_framework}).


\textbf{P4: Offer resource-aware management that saves user expense.} 
\sys implements a resource-aware module and offers advice to users, which can save expense or resources for users especially in the cloud environment.
Using performance-resource extrapolation (Sec. \ref{sec:pr_extra}),
\sys can estimate 1) the minimal number of workers users need to complete the current task within the given time budget, or 2) the minimal time budget to finish the current task given a fixed number of workers.
For tasks that involve expensive-to-evaluate functions, low-fidelity or early-stopped evaluations with less cost could help accelerate the convergence of the optimization process (Sec. ~\ref{sec:add_opt}). 

\begin{figure}[t]
    \centering
    \begin{minipage}[t]{0.95\linewidth}
    \tiny
    \begin{lstlisting}
task_config = {
  "parameter": {
    "x1": { "type": "float", "default": 0, 
      "bound": [-5, 10]},
    "x2": {"type": "int", "bound": [0, 15]},
    "x3": {"type": "cat", "default": "a1", 
      "choice": ["a1", "a2", "a3"]},
    "x4": {"type": "ord", "default": 1, 
      "choice": [1, 2, 3]}},
  "condition": {
      "cdn1": {"type": "equal", "parent": "x3",
        "child": "x1", "value": "a3"}},
  "number_of_trials": 200,
  "time_budget": 10800,
  "task_type": "soc",
  "parallel_strategy": "async",
  "worker_num": 10,
  "use_history": True
  }

    \end{lstlisting}
    \end{minipage}
    \caption{An example of Task Description Language.}
    \label{fig:tdl}
\end{figure}

\subsection{System Architecture}
Based on these design principles, we build \sys as depicted in Figure~\ref{fig:sys_architecture}, which includes five main components.
\textit{Service Master} is responsible for node management, load balance, and fault tolerance.
\textit{Task Database} holds the states of all tasks.
\textit{Suggestion Service} creates new configurations for each task.
\textit{REST API} establishes the bridge between users/workers and suggestion service. \textit{Evaluation workers} are provided and owned by the users. 

\section{System Design}
In this section, we elaborate on the main features and components of \sys from a service perspective.

\subsection{Service Interfaces}
\subsubsection{Task Description Language} For ease of usage, we design a Task Description Language (TDL) to define the optimization task. 
The essential part of TDL is to define the search space, which includes the type and bound for each parameter and the relationships among them.
The parameter types --- \texttt{FLOAT}, \texttt{INTEGER}, \texttt{ORDINAL} and \texttt{CATEGORICAL} are supported in \sys.
In addition, users can add conditions of the parameters to further restrict the search space. Users can also specify the time budget, task type, number of workers, parallel strategy and use of history in TDL. 
Figure~\ref{fig:tdl} gives an example of TDL. 
It defines four parameters \texttt{x1-4} of different types and a condition \texttt{cdn1}, which indicates that \texttt{x1} is active only if \texttt{x3 = ``a3''}. The time budget is three hours, the parallel strategy is \texttt{async}, and transfer learning is enabled.

\subsubsection{Basic Workflow}
Given the TDL for a task, the basic workflow of \sys is implemented as follows:

\begin{lstlisting}[language=python]
# Register the worker with a task.
global_task_id = worker.CreateTask(task_tdl)
worker.BindTask(global_task_id) 
while not worker.TaskFinished():
    # Obtain a configuration to evaluate.
    config = worker.GetSuggestions()
    # Evaluate the objective function. 
    result = Evaluate(config)
    # Report the evaluated results to the server.
    worker.UpdateObservations(config, result)
\end{lstlisting}
Here \texttt{Evaluate} is the evaluation procedure of objective function provided by users. 
By calling \texttt{CreateTask}, the worker obtains a globally unique identifier \texttt{global\_task\_id}. All workers registered with the same \texttt{global\_task\_id} are guaranteed to link with the same task, which enables parallel evaluations. 
While the task is not finished, the worker continues to call \texttt{GetSuggestions} and \texttt{UpdateObservations} to pull suggestions from the suggestion service and update their corresponding observations.

\subsubsection{Interfaces} Users can interact with the \sys service via a \texttt{REST API}. We list the most important service calls as follows:
\begin{itemize}
    \item \texttt{Register}: It takes as input the \texttt{global\_task\_id}, which is created when calling \texttt{CreateTask} from workers, and binds the current worker with the corresponding task. This allows for sharing the optimization history across multiple workers.
    \item \texttt{Suggest}: It suggests the next configurations to evaluate, given the historical observations of the current task.
    \item \texttt{Update}: This method updates the optimization history with the observations obtained from workers. The observations include three parts: the values of the objectives, the results of constraints, and the evaluation information.
    \item \texttt{StopEarly}: It returns a boolean value that indicates whether the current evaluation should be stopped early.
    \item \texttt{Extrapolate}: It uses performance-resource extrapolation, and interactively gives resource-aware advice to users.
\end{itemize}

\subsection{Automatic Algorithm Selection}
\sys implements a wide range of optimization algorithms to achieve high performance in various BBO problems. 
Unlike the existing software packages that use the same algorithm for each task and the same setting for each algorithm, \sys chooses the proper algorithm and setting according to the characteristic of the incoming task. 
We use the classic EI~\cite{movckus1975bayesian} for single-objective optimization task. For multi-objective problems, we select EHVI~\cite{Emmerich2005} when the number of objectives is less than 5; we use MESMO~\cite{Belakaria2019} algorithm for problems with a larger number of objectives, since EHVI's complexity increases exponentially as the number of objectives increases, which not only incurs a large computational overhead but also accumulates floating-point errors. 
We select the surrogate models in BO depending on the configuration space and the number of trials: If the input space has conditions, such as one parameter must be less than another parameter, or there are over 50 parameters in the input space, or the number of trials exceeds 500, we choose the Probabilistic Random Forest proposed in~\cite{hutter2011sequential} instead of Gaussian Process (GP) as the surrogate to avoid incompatibility or high computational complexity of GP. Otherwise, we use GP~\cite{eggensperger2015efficient}.
In addition, \sys will use the \textit{L-BFGS-B} algorithm to optimize the acquisition function if the search space only contains \texttt{FLOAT} and \texttt{INTEGER} parameters; it applies an interleaved local and random search when some of the parameters are not numerical. 
More details about the algorithms implemented in \sys are discussed in Appendix A.2. 

\subsection{Parallel Infrastructure}
\sys is designed to generate suggestions for a large number of tasks concurrently, and a single machine would be insufficient to handle the workload. 
Our suggestion service is therefore deployed across several machines, called \textit{suggestion servers}. 
Each \textit{suggestion server} generates suggestions for several tasks in parallel, giving us a massively scalable suggestion infrastructure. 
Another main component is \textit{service master}, which is responsible for managing the \textit{suggestion servers} and balancing the workload.
It serves as the unified endpoint, and accepts the requests from workers; in this way, each worker does not need to know the dispatching details.
The worker requests new configurations from the \textit{suggestion server} and the \textit{suggestion server} generates these configurations based on an algorithm determined by the automatic algorithm selection module.
Concretely, in this process, the suggestion server utilizes the local penalization based parallelization mechanism (Sec. ~\ref{lp_based_parallel}) and transfer-learning framework (Sec. ~\ref{tl_framework}) to improve the sample efficiency.

One main design consideration is to maintain a fault-tolerant production system, as machine crash happens inevitably.
In \sys, the \textit{service master} monitors the status of each server and preserves a table of active servers.
When a new task comes, the \textit{service master} will assign it to an active server and record this binding information.
If one server is down, its tasks will be dispatched to a new server by the master, along with the related optimization history stored in the task database.
Load balance is one of the most important guidelines to make such task assignments.
In addition, the snapshot of \textit{service master} is stored 
in the remote database service; if the master is down, we can recover it by restarting the node and fetching the snapshot from the database. 

\subsection{Performance-Resource Extrapolation}
\label{sec:pr_extra}

In the setting of parallel infrastructure with cloud computing, saving expense is one of the most important concerns from users.
\sys can guide users to configure their resources, e.g., the minimal number of workers or time budget, which further saves expense for users.
Concretely, we use a weighted cost model to extrapolate the performance vs. trial curve. It uses several parametric decreasing saturating function families as base models, and we apply MCMC inference to estimate the parameters of the model. 
Given the existing observations, \sys trains a cost model as above and uses it to predict the number of trials at which the curve approaches the optimum.
Based on this prediction and the cost of each evaluation, \sys estimates the minimal resource needed to reach satisfactory performance (more details in Appendix A.1).

\vspace{-0.5em}
\paragraph*{\underline{Application Example}} Two interesting applications that save expense for users are listed as follows:

\noindent
\textit{Case 1.}
Given a fixed number of workers, \sys outputs a minimal time budget $B_{\text{min}}$ to finish this task based on the estimated evaluation cost of workers. 
With this estimation, users can stop the task in advance if the given time budget $B_{\text{task}} > B_{\text{min}}$; otherwise, users should increase the time budget to $B_{\text{min}}$.

\noindent
\textit{Case 2.}
Given a fixed time budget $B_{\text{task}}$ and initial number of workers, \sys can suggest the minimal number of workers $N_{\text{min}}$ to finish the current task within $B_{\text{task}}$ by adjusting the number of workers to $N_{\text{min}}$ dynamically.

\subsection{Augmented Components in \sys}

\textit{\underline{Extensibility and Benchmark Support.}} 
\sys's modular design allows users to define their suggestion algorithms easily by inheriting and implementing an abstract \texttt{Advisor}. 
The key abstraction method of \texttt{Advisor} is \texttt{GetSuggestions}, which receives the observations of the current task and suggests the next configurations to evaluate based on the user-defined policy.
In addition, \sys provides a \textit{benchmark suite} of various BBO problems to benchmark the optimization algorithms. 

\noindent
\textit{\underline{Data Privacy Protection.}} 
In some scenarios, the names and ranges of parameters are sensitive, e.g., in hyper-parameter tuning, the parameter names may reveal the architecture details of neural networks. 
To protect data privacy, the \texttt{REST API} applies a transformation to anonymize the parameter-related information before sending it to the service. 
This transformation involves 1) converting the parameter names to some regular ones like ``param1'' and 2) rescaling each parameter to a default range that has no semantic. 
The workers can perform an inverse transformation when receiving an anonymous configuration from the service.  

\begin{figure}[t]
\centering
\includegraphics[width=0.97\columnwidth]{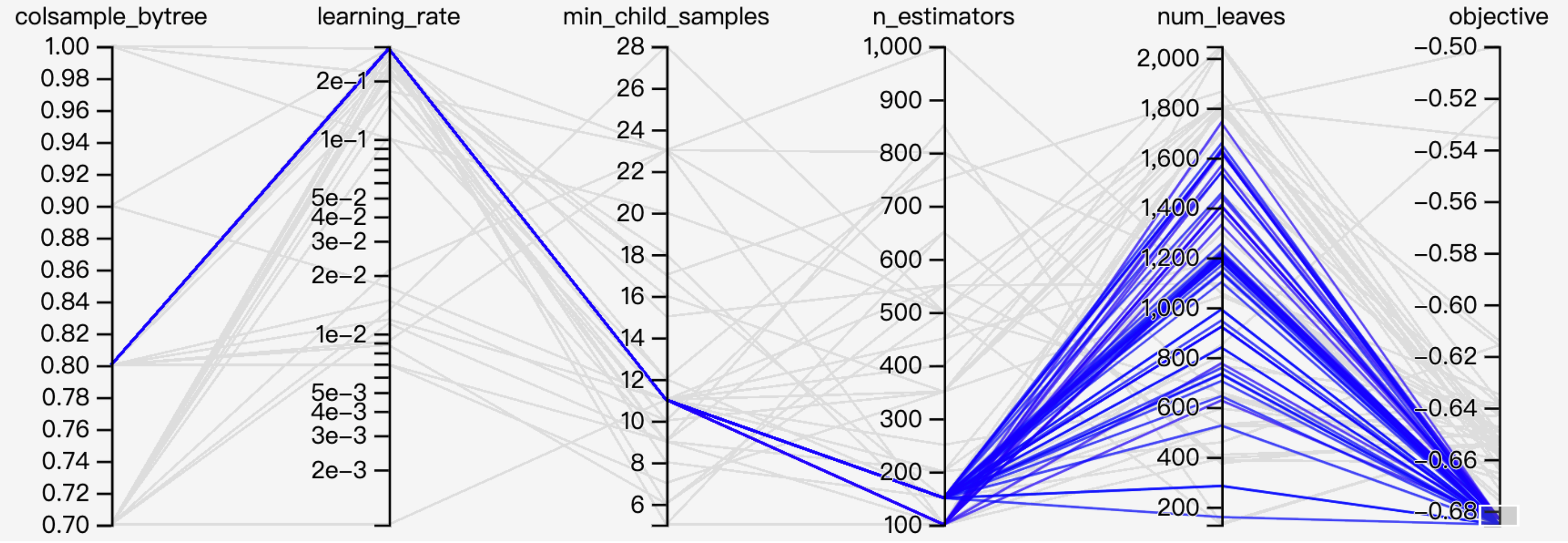}
\vspace{-1em}
\caption{An example of the Parallel Coordinates Visualization for configurations when tuning \textit{LightGBM}.}
\label{fig:visualization}
\end{figure}

\noindent
\textit{\underline{Visualization.}} 
\sys provides an online dashboard based on \texttt{TensorBoardX} which enables users to monitor the optimization process and check the evaluation info of the current task.
Figure \ref{fig:visualization} visualizes the evaluation results in a hyper-parameter tuning task.

\section{System Optimizations}

\subsection{Local Penalization based Parallelization}
\label{lp_based_parallel}
\begin{figure}[t]
\centering
\includegraphics[width=0.48\textwidth]{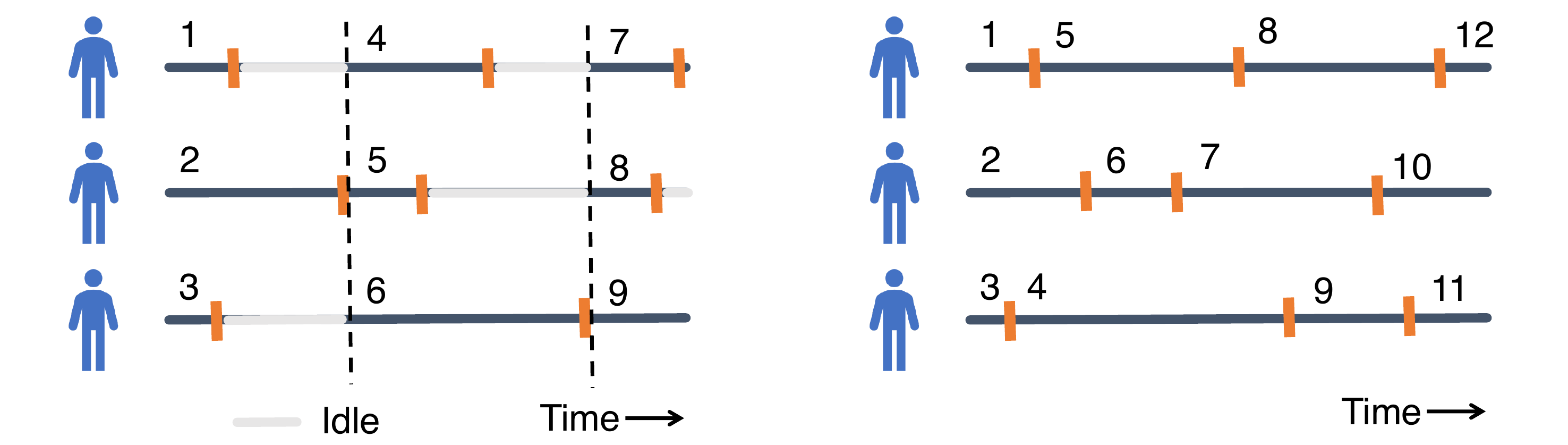}
\vspace{-2em}
\caption{An illustration of the synchronous (left) and asynchronous (right) parallel methods using three workers. The numbers above the horizontal lines are the configuration ids, and the short vertical lines indicate when a worker finished the evaluation of last configuration.}
\label{fig:parallel_bo}
\end{figure}

Most proposed Bayesian optimization (BO) approaches only allow the exploration of the parameter space to occur sequentially.
To fully utilize the computing resources in a parallel infrastructure, we provide a mechanism for distributed parallelization, where multiple configurations can be evaluated concurrently across workers. 
Two parallel settings are considered (see Figure~\ref{fig:parallel_bo}):

\noindent
\underline{\emph{1) Synchronous parallel setting.}}
The worker pulls new configuration from \textit{suggestion server} to evaluate until all the workers have finished their last evaluations.

\noindent
\underline{\emph{2) Asynchronous parallel setting.}}
The worker pulls a new configuration when the previous evaluation is completed.

Our main concern is to design an algorithm-agnostic mechanism that can parallelize the optimization algorithms under the sync and async settings easily, so we do not need to implement the parallel version for each algorithm individually.
To this end, we propose a local penalization based parallelization mechanism, the goal of which is to sample new configurations that are promising and far enough from the configurations being evaluated by other workers. 
This mechanism can handle the well-celebrated exploration vs. exploitation trade-off, and meanwhile prevent workers from exploring similar configurations.
Algorithm~\ref{algo:paralllel_sample} gives the pseudo-code of sampling a new configuration under the sync/async settings.
More discussion about this is provided in Appendix A.4.

\subsection{General Transfer-Learning Framework}
\label{tl_framework}
When performing BBO, users often run tasks that are similar to previous ones. This fact can be used to speed up the current task. 
Compared with \texttt{Vizier}, which only provides limited transfer learning functionality for single-objective BBO problems, \sys employs a general transfer learning framework with the following advantages: 1) support for the generalized black-box optimization problems, and 2) compatibility with most BO methods.

\sys takes as input observations from $K+1$ tasks: $D^1$, ..., $D^K$
for $K$ previous tasks and $D^T$ for the current task.
Each $D^i=\{(\bm{x}_j^i, \bm{y}_j^i)\}_{j=1}^{n_i},\ i=1,...,K $, includes a set of observations.
Note that, $\bm{y}$ is an array, including multiple objectives for configuration $\bm{x}$.

For multi-objective problems with $p$ objectives, we propose to transfer the knowledge about $p$ objectives individually. 
Thus, the transfer learning of multiple objectives is turned into $p$ single-objective transfer learning processes.
For each dimension of the objectives, we take RGPE~\cite{feurer2018scalable} as the base method.
1) We first train a surrogate model $M^i$ on $D^i$ for the $i^{th}$ prior task and $M^T$ on $D^T$; based on $M^{1:K}$ and $M^{T}$, 2) we then build a transfer learning surrogate by combining all base surrogates:
\[
M^{\text{TL}} = \texttt{agg}(\{M^1,...,M^K, M^{T}\}; {\bf w});
\]
3) the surrogate $M^{\text{TL}}$ is used to guide the configuration search, instead of the original $M^{T}$. 
Concretely, we combine the multiple base surrogates ($\texttt{agg}$) linearly, and the parameters {\bf w} are calculated based on the ranking of configurations, which reflects the similarity between the source and target task (see details in Appendix A.3).

\para{Scalability discussion} A more intuitive alternative is to obtain a transfer learning surrogate by using all observations from $K+1$ tasks, and this incurs a complexity of $\mathcal{O}(k^3n^3)$ for $k$ tasks with $n$ trials each (since GP has $\mathcal{O}(n^3)$ complexity). 
Therefore, it is hard to scale to a larger number of source tasks (a large $k$). 
By training base surrogates individually, the proposed framework is a more computation-efficient solution that has $\mathcal{O}(kn^3)$ complexity.

\begin{algorithm}[t]
  \scriptsize
  \SetAlgoLined
  \caption{Pseudo code for \emph{Sample} configuration}
  \KwIn{the hyper-parameter space $\mathcal{X}$, configuration observations $D=\{(\bm{x_i},\bm{y_i})\}_{i=1}^n$, configurations being evaluated $C_{\text{eval}}$, surrogate model $M$, and acquisition function $\alpha(\cdot)$.}
  \SetAlgoLined
  \label{algo:paralllel_sample}
  calculate $\hat{\bm{y}}$, the median of $\{\bm{y_i}\}_{i=1}^n$\;
  create new observations $D_{\text{new}}=\{(\bm{x}_\text{eval}, \hat{\bm{y}}): \bm{x}_\text{eval} \in C_{\text{eval}}\}$\;
  fit a surrogate model $M$ (e.g., a GP) on $D_{\text{aug}}$, where $D_{\text{aug}} = D \cup D_{\text{new}}$, and build the acquisition function $\alpha(\bm{x}, M)$ using $M$\;
  \textbf{return} the configuration $\bar{\bm{x}}=\operatorname{argmax}_{\bm{x} \in \mathcal{X}}\alpha(\bm{x}, M)$.
\end{algorithm}

\begin{figure*}[t]
	\centering
	\subfigure[2d-Branin]{
		\scalebox{0.23}[0.23]{
			\includegraphics[width=1\linewidth]{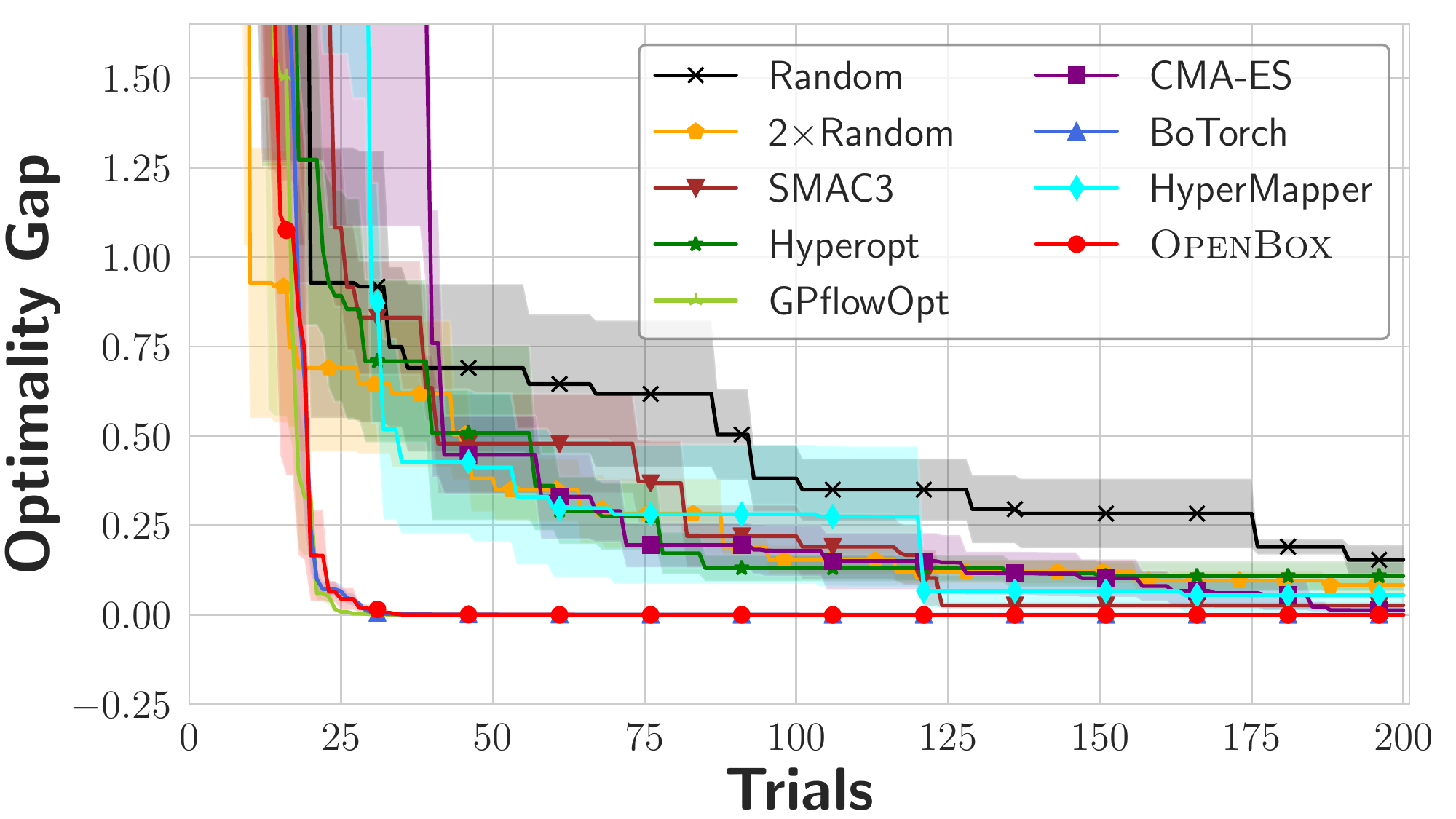}
			\label{so_branin}
	}}
	\subfigure[2d-Ackley]{
		\scalebox{0.23}[0.23]{
			\includegraphics[width=1\linewidth]{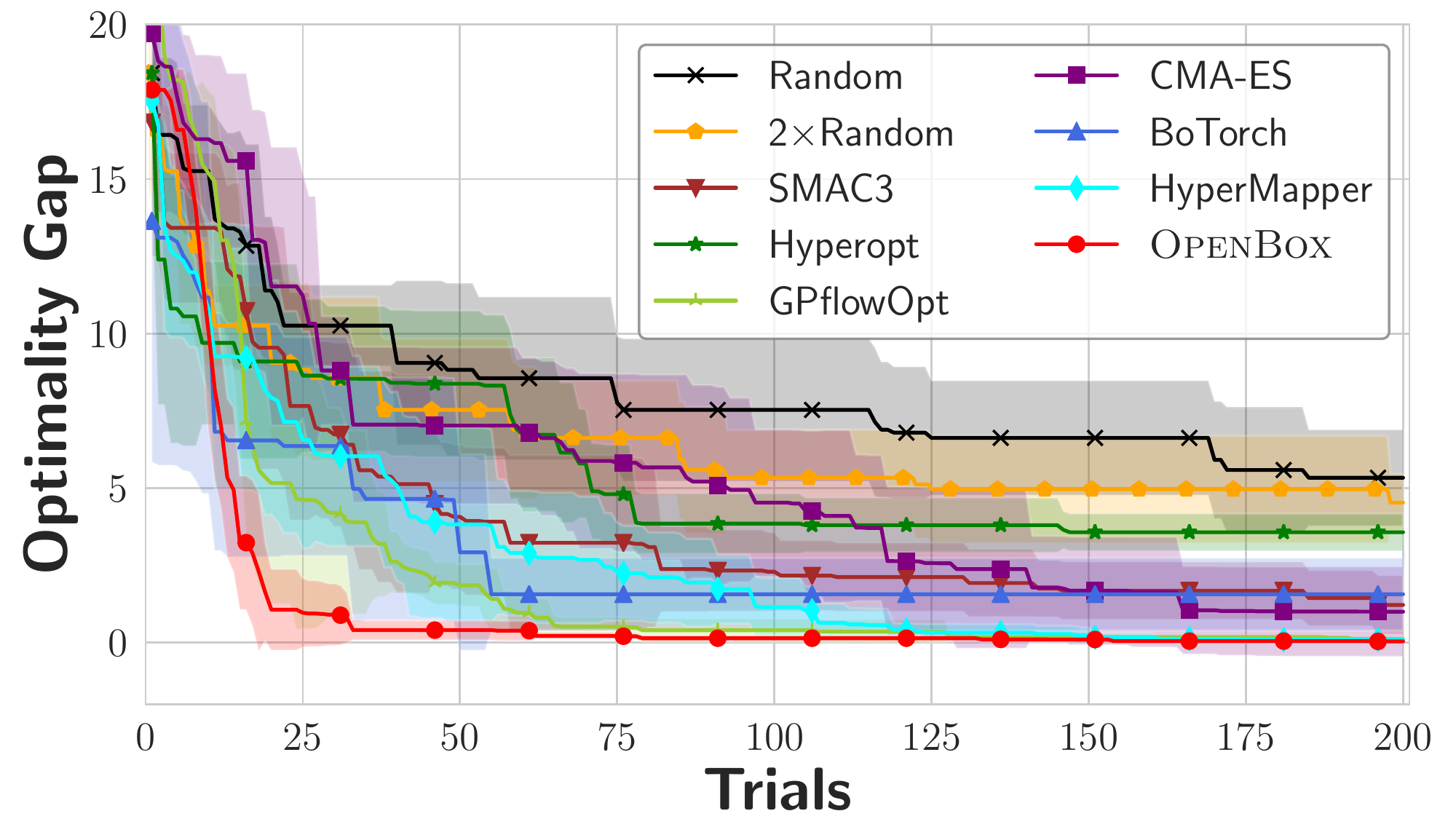}
			\label{so_ackley}
	}}
	\subfigure[2d-Beale]{
		\scalebox{0.23}[0.23]{
			\includegraphics[width=1\linewidth]{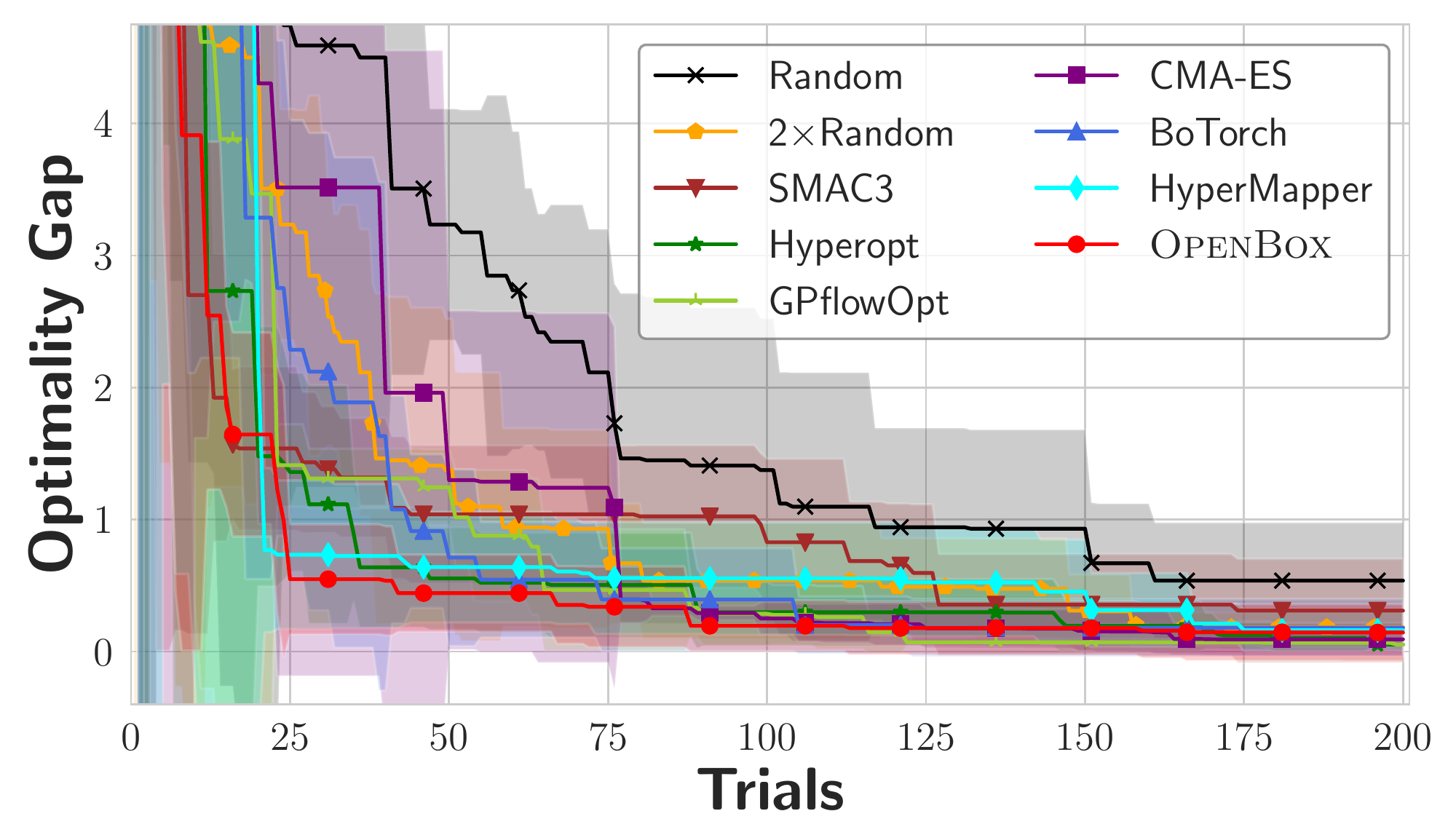}
			\label{so_beale}
	}}
	\subfigure[6d-Hartmann]{
		\scalebox{0.23}[0.23]{
			\includegraphics[width=1\linewidth]{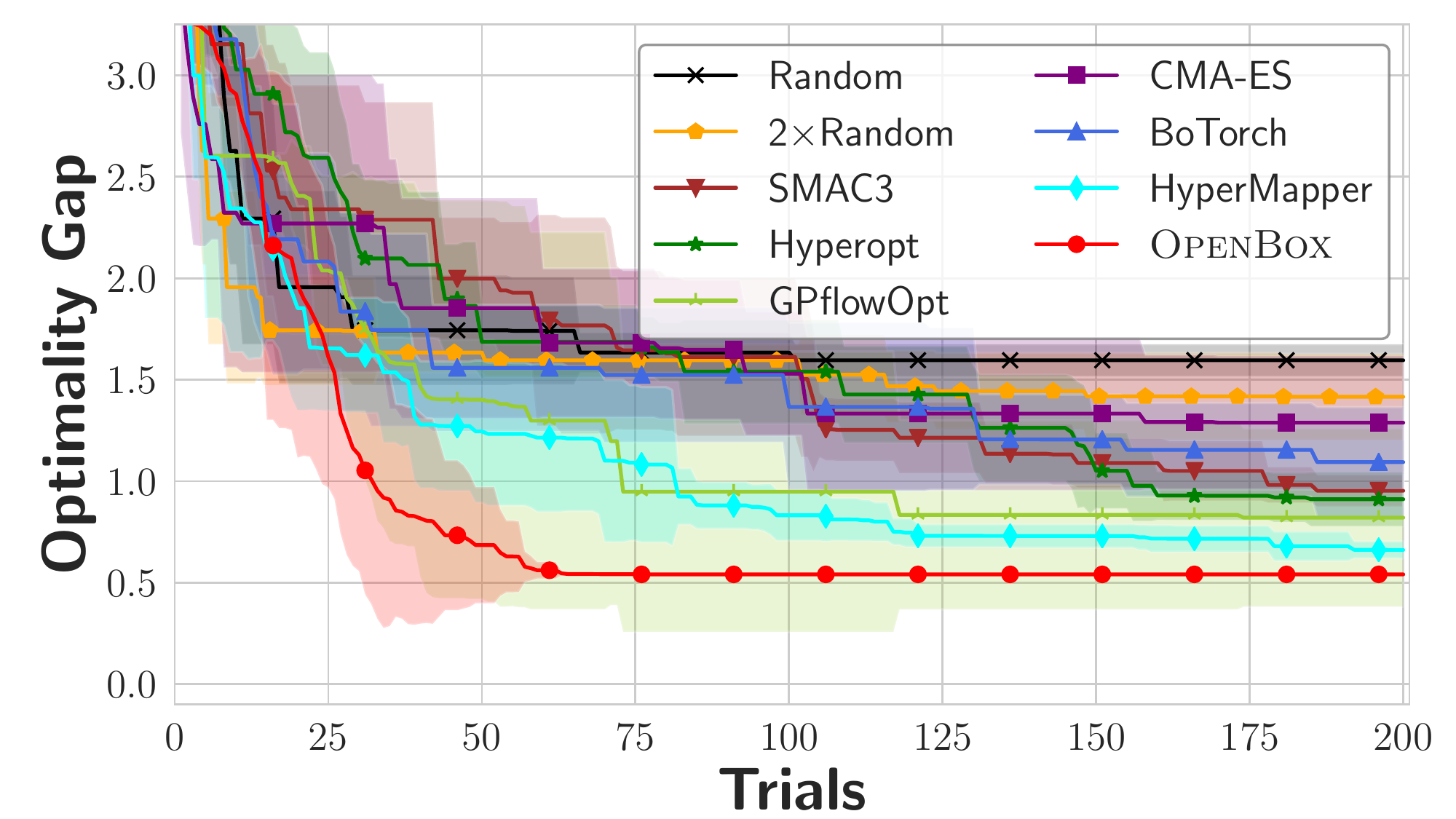}
			\label{so_hartmann}
	}}
	\vspace{-1.5em}
	\caption{Results for four black-box problems with single objective.}
	\vspace{-1.5em}
  \label{fig:sobo}
\end{figure*}

\begin{figure*}[t]
	\centering
	\subfigure[4d-Ackley]{
		\scalebox{0.23}[0.23]{
			\includegraphics[width=1\linewidth]{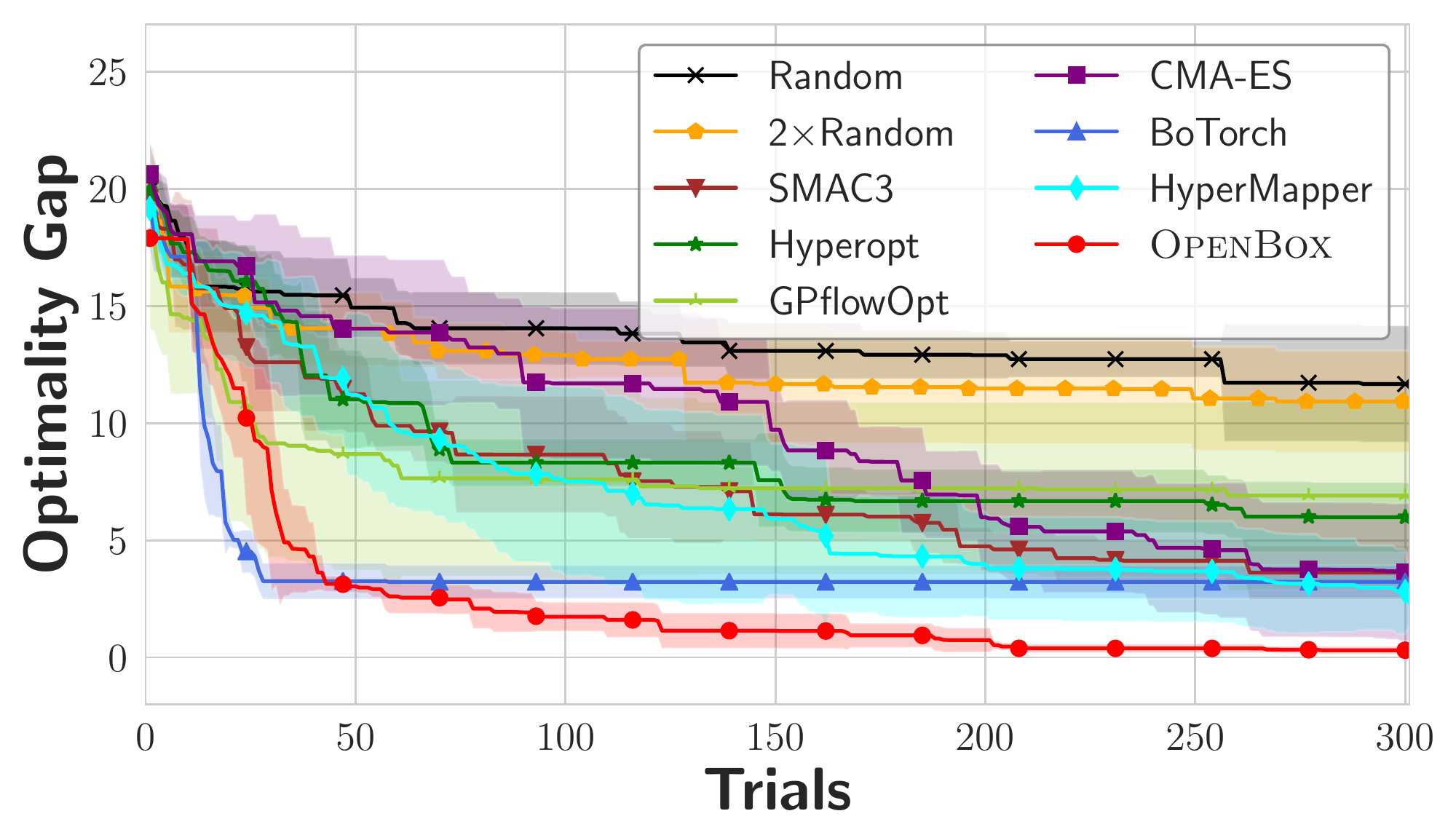}
			\label{ackley-4}
	}}
	\subfigure[8d-Ackley]{
		\scalebox{0.23}[0.23]{
			\includegraphics[width=1\linewidth]{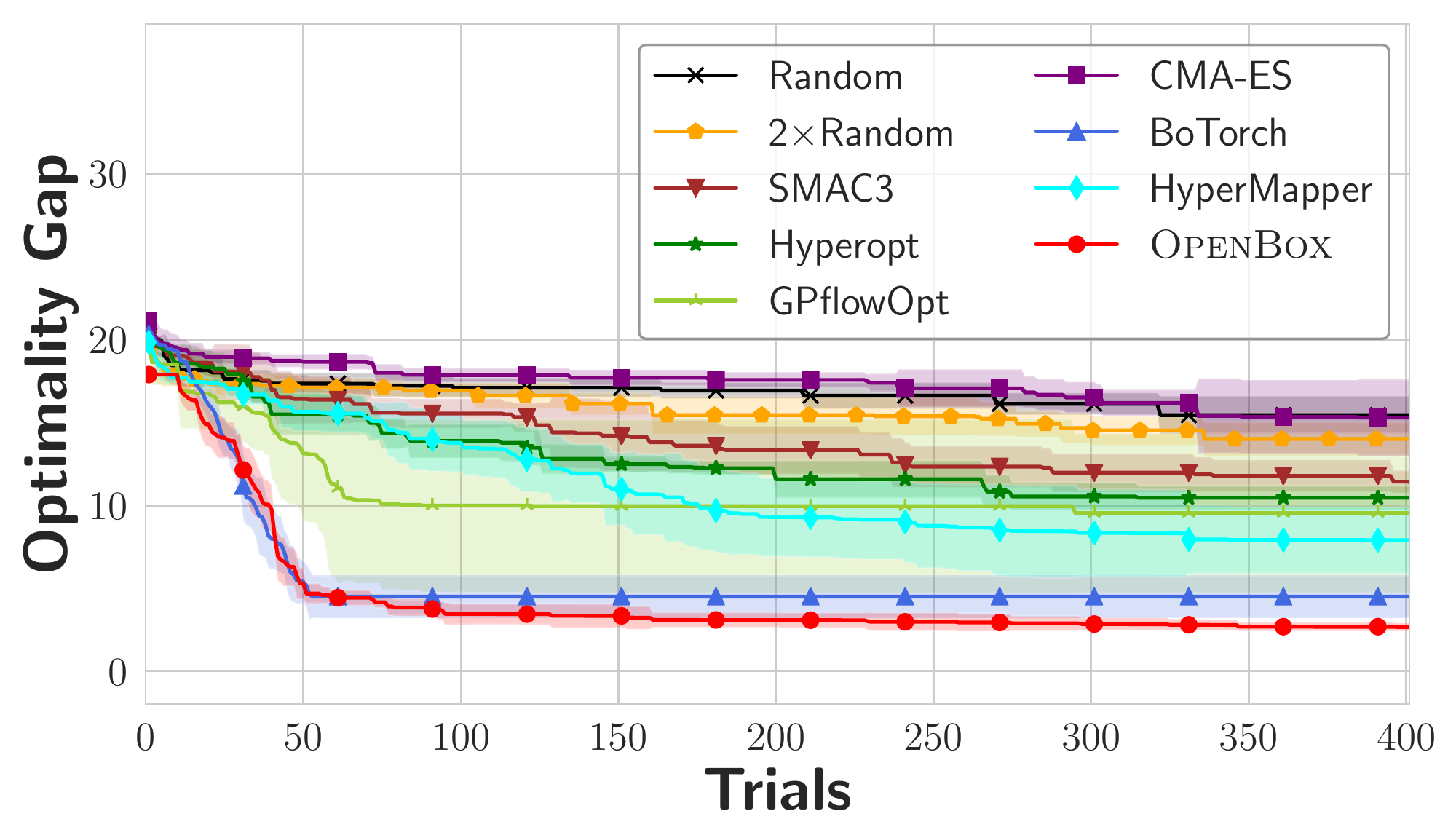}
			\label{ackley-8}
	}}
	\subfigure[16d-Ackley]{
		\scalebox{0.23}[0.23]{
			\includegraphics[width=1\linewidth]{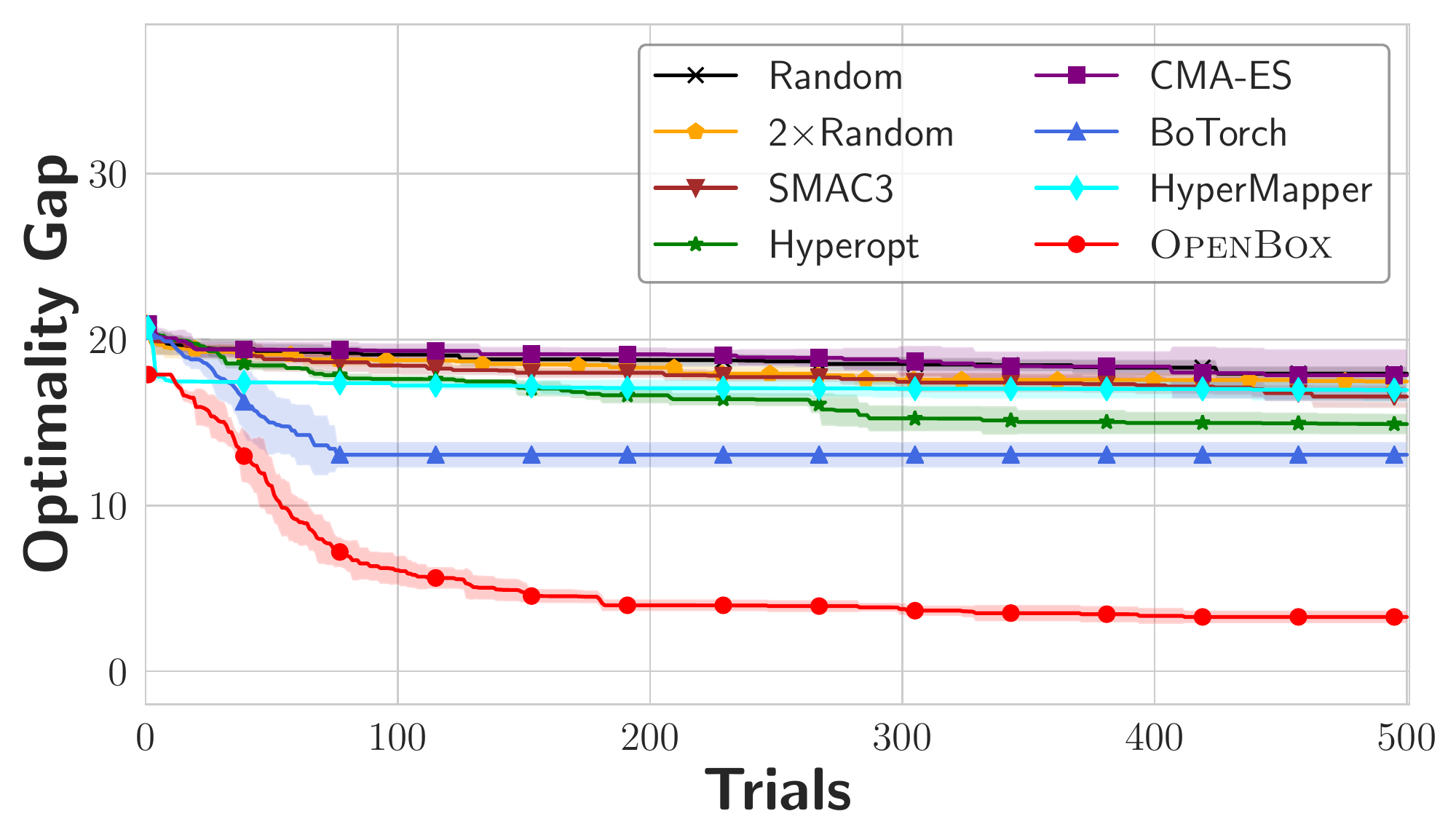}
			\label{ackley-16}
	}}
	\subfigure[32d-Ackley]{
		\scalebox{0.23}[0.23]{
			\includegraphics[width=1\linewidth]{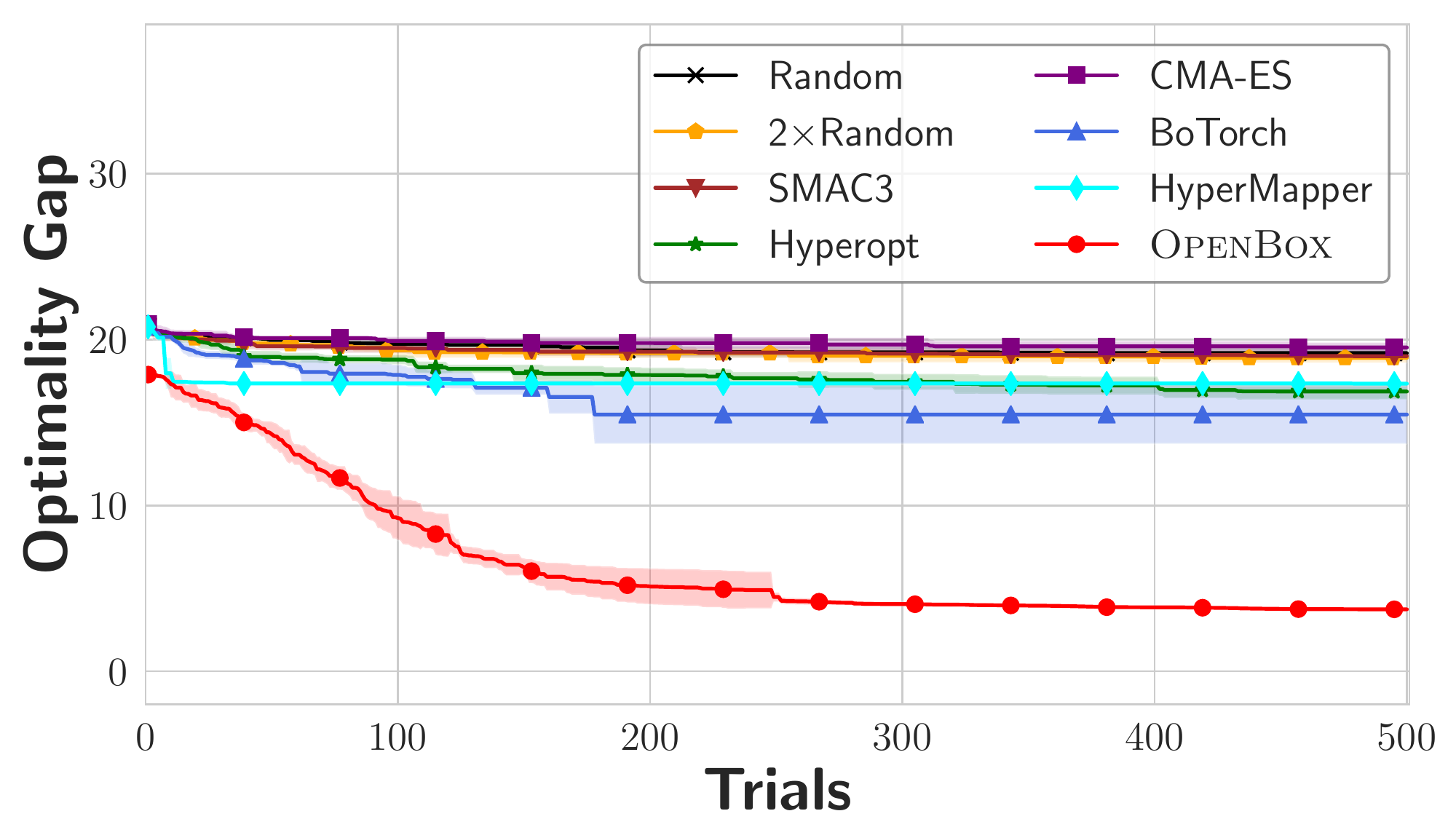}
			\label{ackley-32}
	}}
	\vspace{-1.5em}
	\caption{Scalability results on solving \texttt{Ackley} with different input dimensions.}
	\vspace{-1.5em}
  \label{fig:sobo_scalability}
\end{figure*}

\subsection{Additional Optimizations}
\label{sec:add_opt}
\sys also includes two additional optimizations that can be applied to improve the efficiency of black-box optimizations.

\subsubsection{Multi-Fidelity Support and Applications}
During each evaluation in the multi-fidelity setting~\cite{poloczek2017multi,li2020mfes}, the worker receives an additional parameter, indicating how many resources are used to evaluate this configuration. 
The resource type needs to be specified by users. For example, in hyper-parameter tuning, it can be the number of iterations for an iterative algorithm and the size of dataset subset.
The trial with partial resource returns a low-fidelity result with a cheap evaluation cost. 
Though not as precise as high-fidelity results, the low-fidelity results can provide some useful information to guide the configuration search.
In \sys, we have implemented several multi-fidelity algorithms, such as \texttt{MFES-HB}~\cite{li2020mfes}. 
\subsubsection{Early-Stopping Strategy}
Orthogonal to the above optimization, early-stopping strategies aim to stop a poor trial in advance based on its intermediate results. 
In practice, a worker can periodically ask suggestion service whether it should terminate the current evaluation early.
In \sys, we provide two early-stopping strategies:
1) learning curve extrapolation based methods ~\cite{Domhan2015,Klein2017} that stop the poor configurations by estimating the future performance,
and 2) mean or median termination rules based on comparing the current result with previous ones.

\section{Experimental Evaluation}
In this section, we compare the performance and efficiency of \sys against existing software packages on multiple kinds of black-box optimization tasks, including tuning tasks in AutoML.

\subsection{Experimental Setup}

\subsubsection{Baselines} Besides the systems mentioned in Table~\ref{tbl:sys_cmp}, we also use \texttt{CMA-ES}~\cite{Hansen2001}, \texttt{Random Search} and \texttt{2$\times$Random Search} (Random Search with double budgets) as baselines. 
To evaluate transfer learning, we compare \sys with \texttt{Google Vizier}.
For multi-fidelity experiments, we compare \sys against \texttt{HpBandSter} and \texttt{BOHB}, the details of which are in Appendix A.5.

\subsubsection{Problems} 
We use 12 black-box problems (mathematical functions) from~\cite{Zitzler2000} and two AutoML optimization problems on 25 OpenML datasets. 
In particular, 2d-Branin, 2d-Beale, 6d-Hartmann and (2d, 4d, 8d, 16d, 32d)-Ackley are used for single-objective optimization; 2d-Townsend, 2d-Mishra, 4d-Ackley and 10d-Keane are used for constrained single-objective optimization; 3d-ZDT2 with two objectives and 6d-DTLZ1 with five objectives are used for multi-objective optimization; 2d-CONSTR and 2d-SRN with two objectives are used for constrained multi-objective optimization. 
All the parameters for mathematical problems are of the \texttt{FLOAT} type and the maximum trials of each problem depend on its difficulty, which ranges from 80 to 500. 
For AutoML problems on 25 datasets, we split each dataset and search for the configuration with the best validation performance.
Specifically, we tune \texttt{LightGBM} and \texttt{LibSVM} with the linear kernel, where the parameters of \texttt{LightGBM} are of the \texttt{FLOAT} type while \texttt{LibSVM} contains \texttt{CATEGORICAL} and conditioned parameters.

\subsubsection{Metrics} We employ the three metrics as follows.

\noindent
\textbf{1. Optimality gap} is used for single-objective mathematical problem. That is, if $x^{*}$ optimizes $f$, and $\hat{x}$ is the best configuration found by the method, then $|f(\hat{x})-f(x^{*})|$ measures the success of the method on that function. In rare cases, we report the objective value if the ground-truth optimal $x^*$ is extremely hard to obtain. 

\noindent
\textbf{2. Hypervolume indicator} given a reference point $\bm{r}$ measures the quality of a Pareto front in multi-objective problems. 
We report the \textit{difference} between the hypervolume of the ideal Pareto front $\mathcal{P}^*$ and that of the estimated Pareto front $\mathcal{P}$ by a given algorithm, which is $HV(\mathcal{P}^*, \bm{r}) - HV(\mathcal{P}, \bm{r})$. 

\noindent
\textbf{3. Metric for AutoML.} For single-objective AutoML problems, we report the validation error. To measure the results across different datasets, we use \texttt{Rank} as the metric.

\subsubsection{Parameter Settings} For both \sys and the considered baselines, we use the default setting. Each experiment is repeated 10 times, and we compute the mean and variance for visualization. 

\subsection{Results and Analysis}
\subsubsection{Single-Objective Problems without Constraints}
Figure \ref{fig:sobo} illustrates the results of \sys on different single-objective problems compared with competitive baselines while Figure \ref{fig:sobo_scalability} displays the performance with the growth of input dimensions. 
In particular, Figure \ref{fig:sobo} shows that \sys, \texttt{HyperMapper} and \texttt{BoTorch} are capable of optimizing these low-dimensional functions stably. However, when the dimensions of the parameter space grow larger, as shown in Figure \ref{fig:sobo_scalability}, only \sys achieves consistent and excellent results while the other baselines fail, which demonstrates its scalability on input dimensions.
Note that, \sys achieves more than 10-fold speedups over the baselines when solving \texttt{Ackley} with 16 and 32-dimensional inputs. 

\begin{figure*}[htb]
	\centering
	\subfigure[2d-Townsend]{
		\scalebox{0.23}[0.23]{
			\includegraphics[width=1\linewidth]{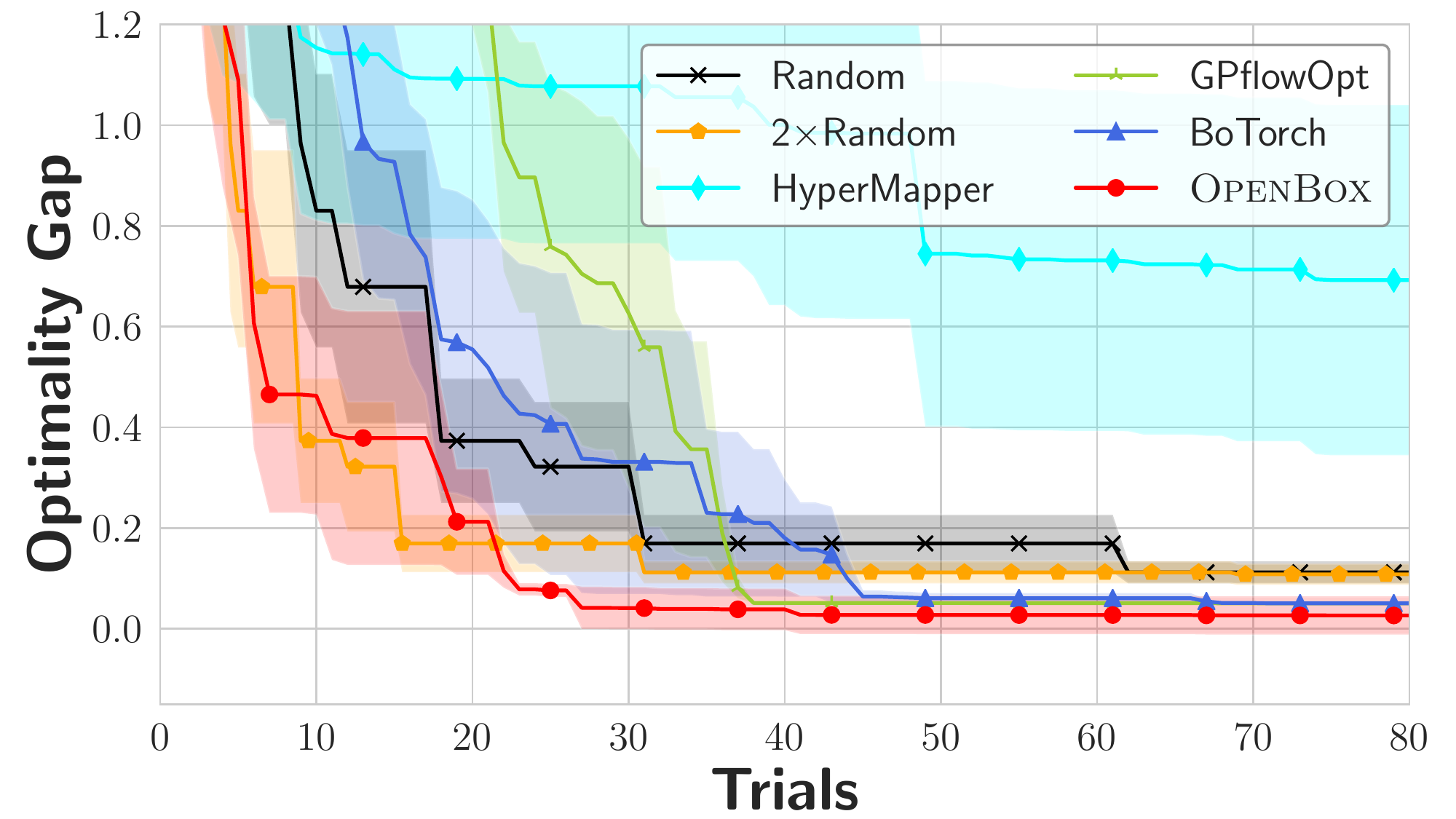}
			\label{soc_townsend}
	}}
	\subfigure[2d-Mishra]{
		\scalebox{0.23}[0.23]{
			\includegraphics[width=1\linewidth]{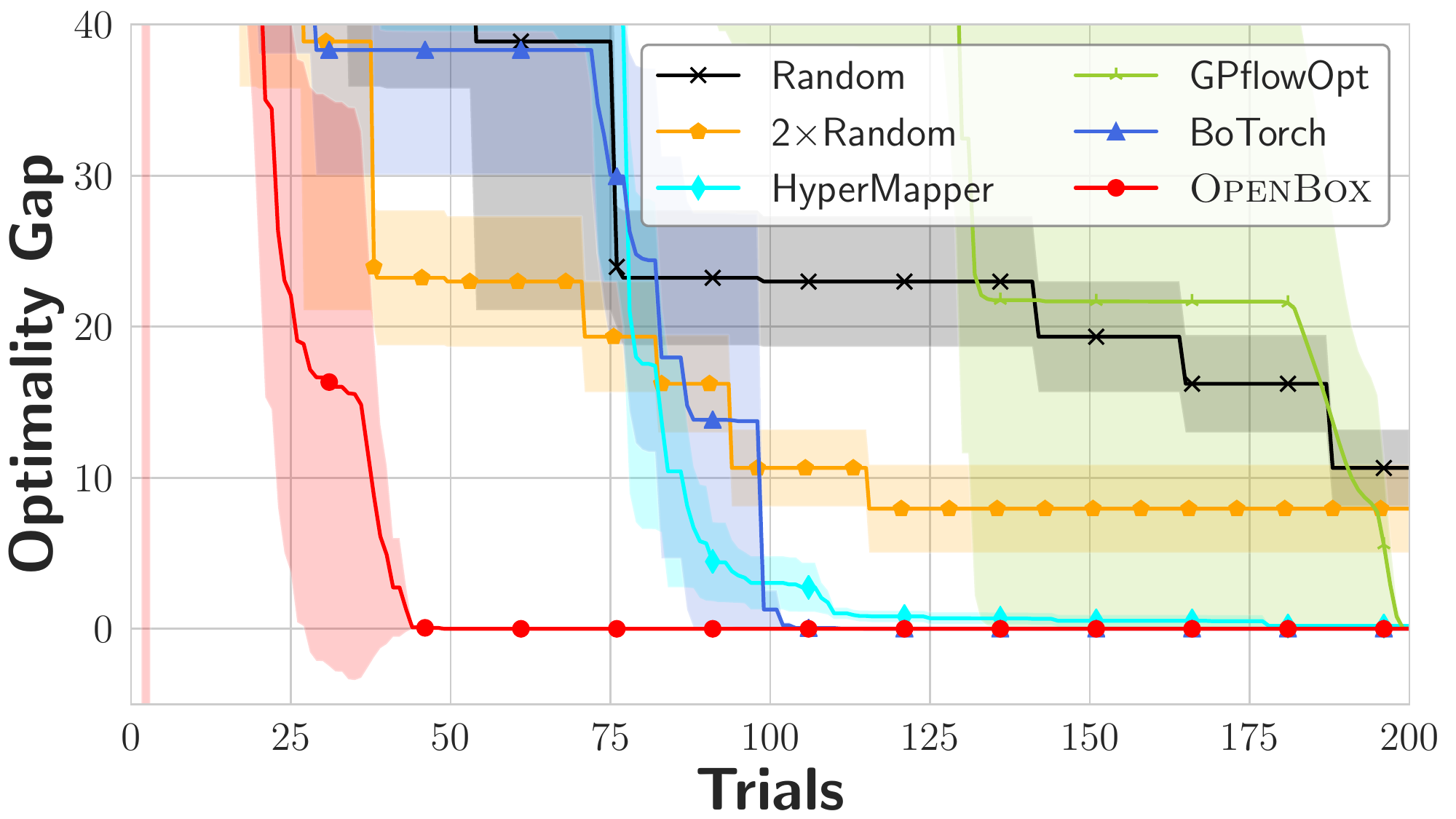}
			\label{soc_mishra}
	}}
	\subfigure[4d-Ackley]{
		\scalebox{0.23}[0.23]{
			\includegraphics[width=1\linewidth]{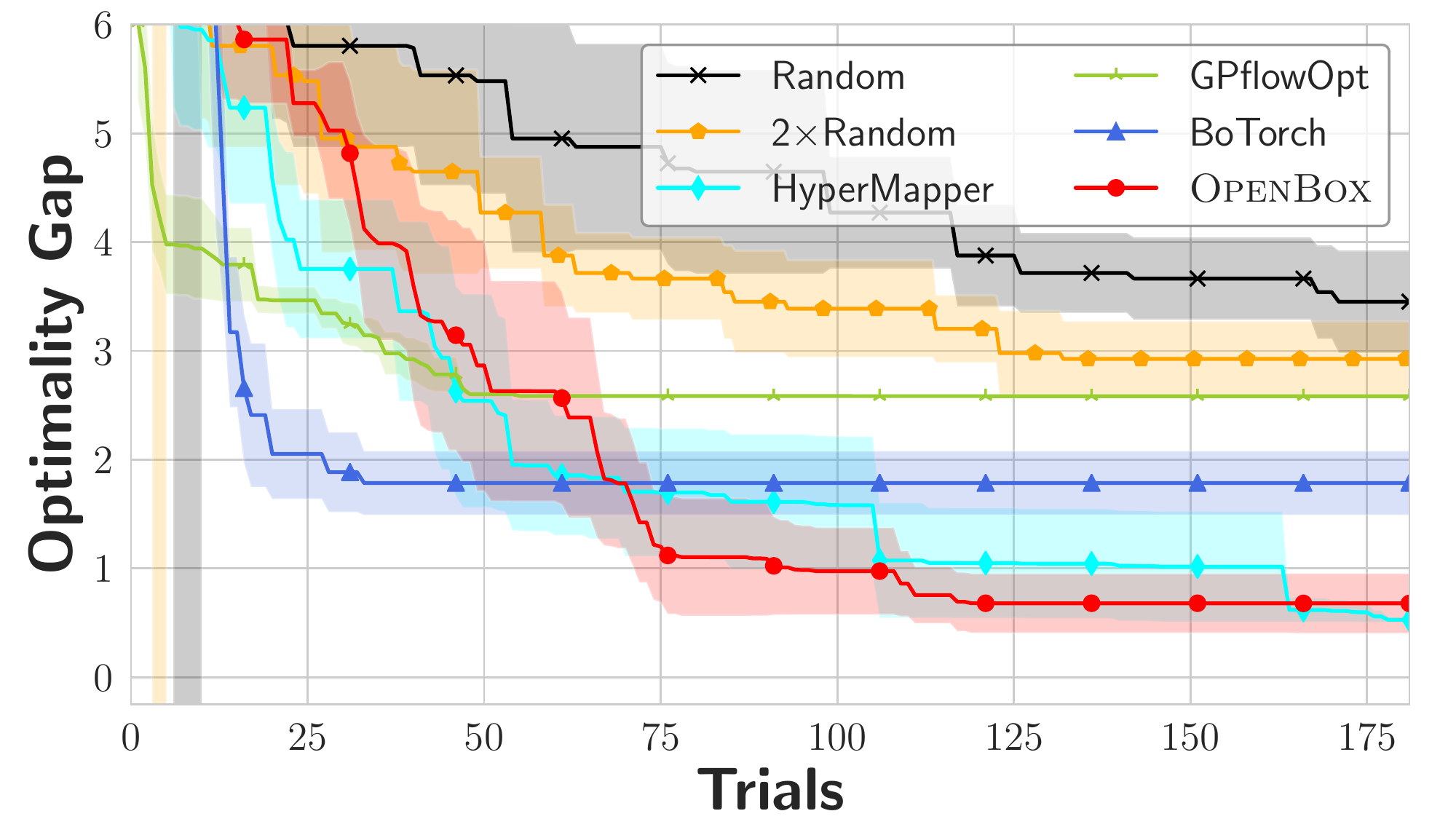}
			\label{soc_ackley}
	}}
	\subfigure[10d-Keane]{
		\scalebox{0.23}[0.23]{
			\includegraphics[width=1\linewidth]{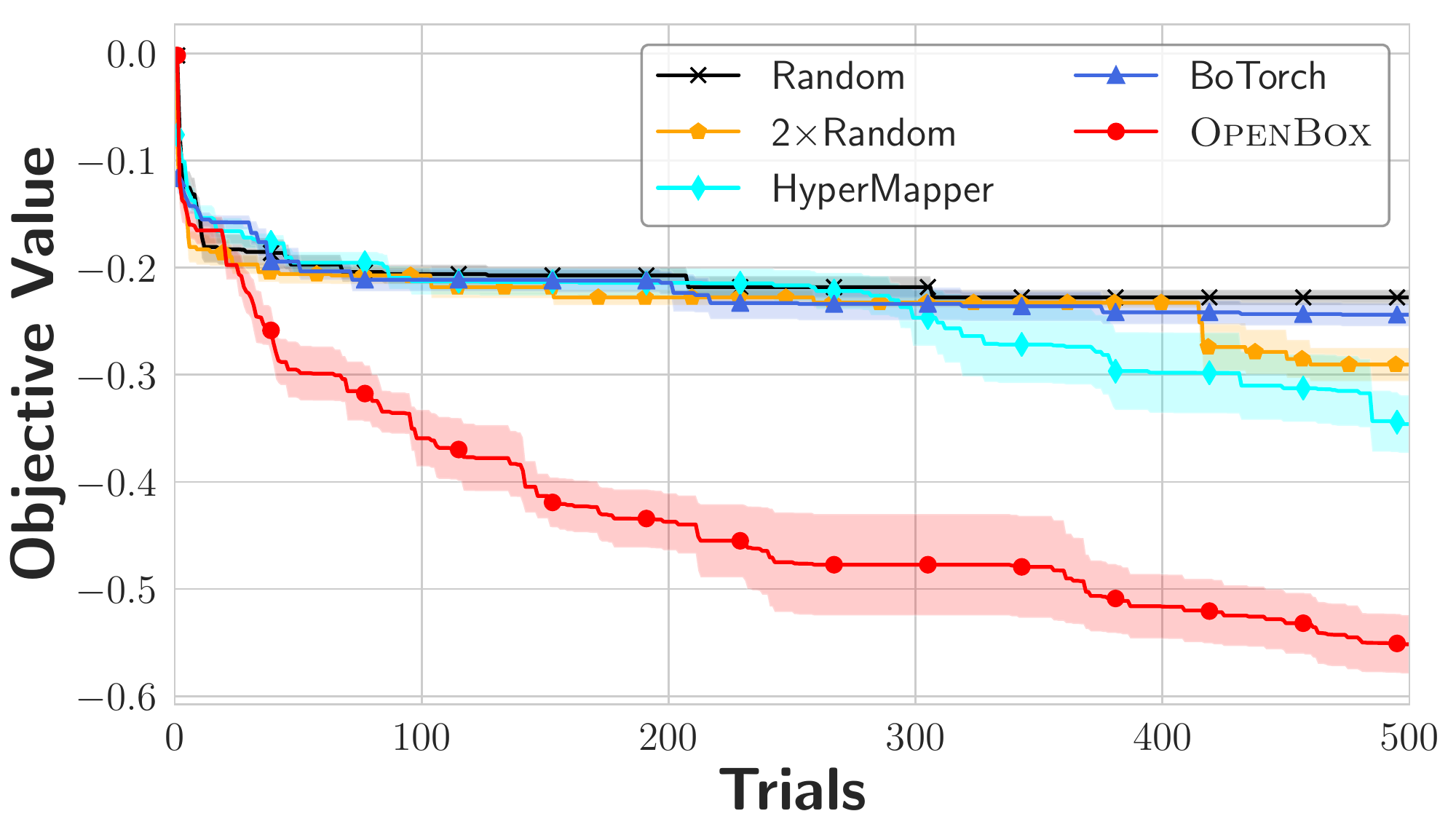}
			\label{soc_keane}
	}}
	\vspace{-1.5em}
	\caption{Results for solving four single-objective black-box problems with constraints.}
	\vspace{-1.5em}
  \label{fig:soboc}
\end{figure*}

\begin{figure*}[htb]
	\centering
	\subfigure[3d-ZDT2]{
		\scalebox{0.23}[0.23]{
			\includegraphics[width=1\linewidth]{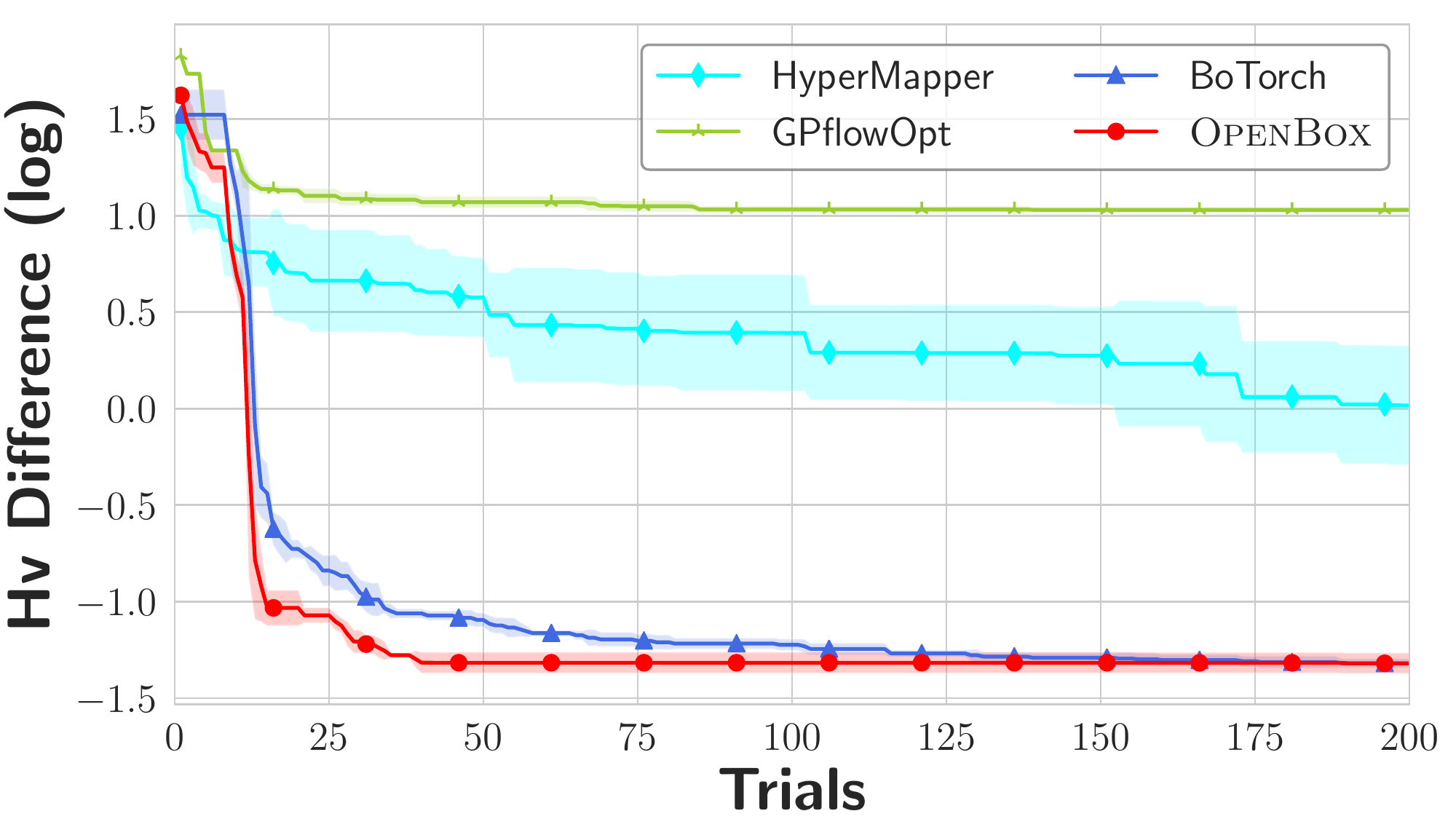}
			\label{mo_zdt2}
	}}
	\subfigure[6d-DTLZ1]{
		\scalebox{0.23}[0.23]{
			\includegraphics[width=1\linewidth]{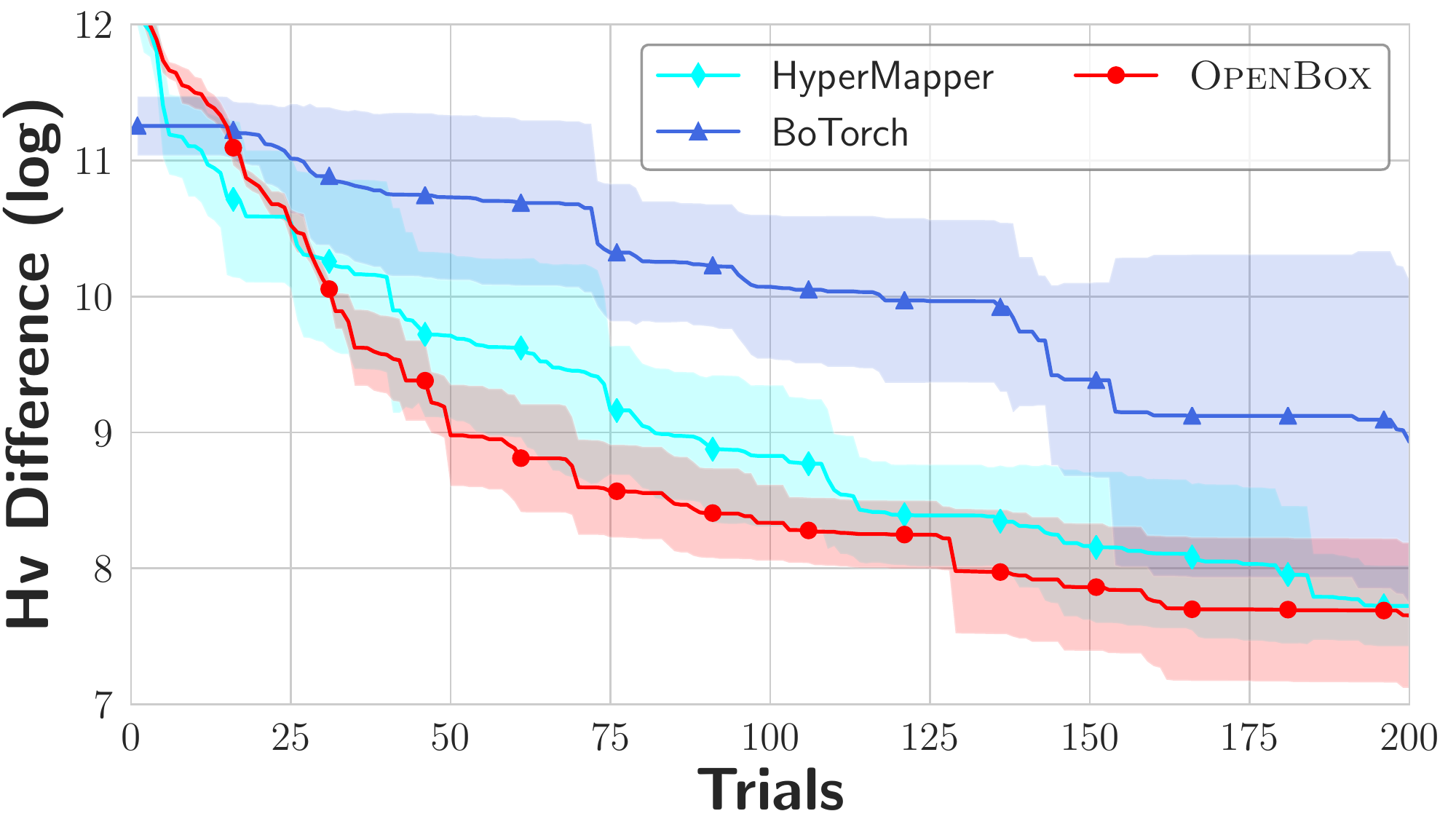}
			\label{mo_dtlz1}
	}}
	\subfigure[2d-CONSTR]{
		\scalebox{0.23}[0.23]{
			\includegraphics[width=1\linewidth]{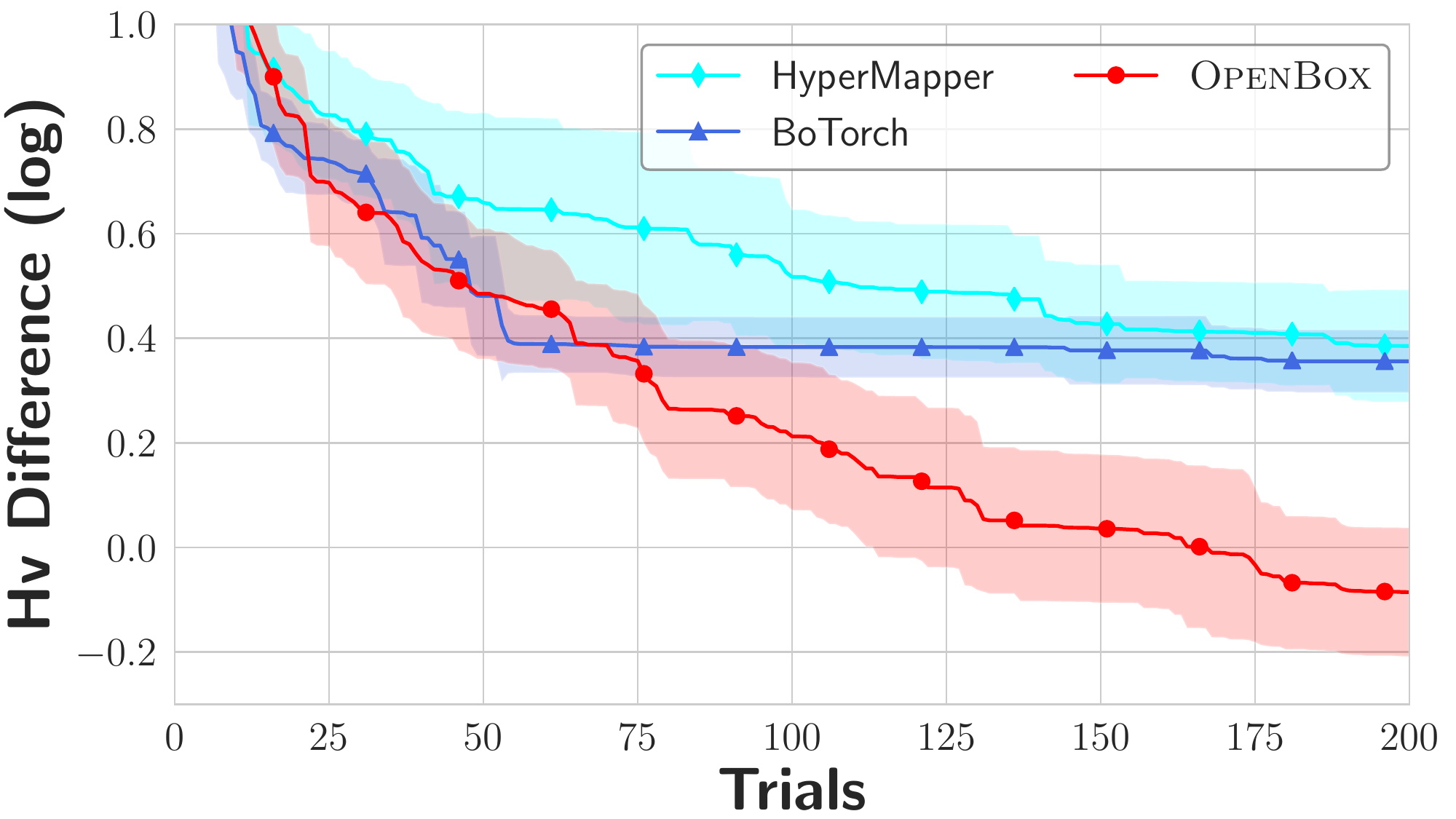}
			\label{moc_constr}
	}}
	\subfigure[2d-SRN]{
		\scalebox{0.23}[0.23]{
			\includegraphics[width=1\linewidth]{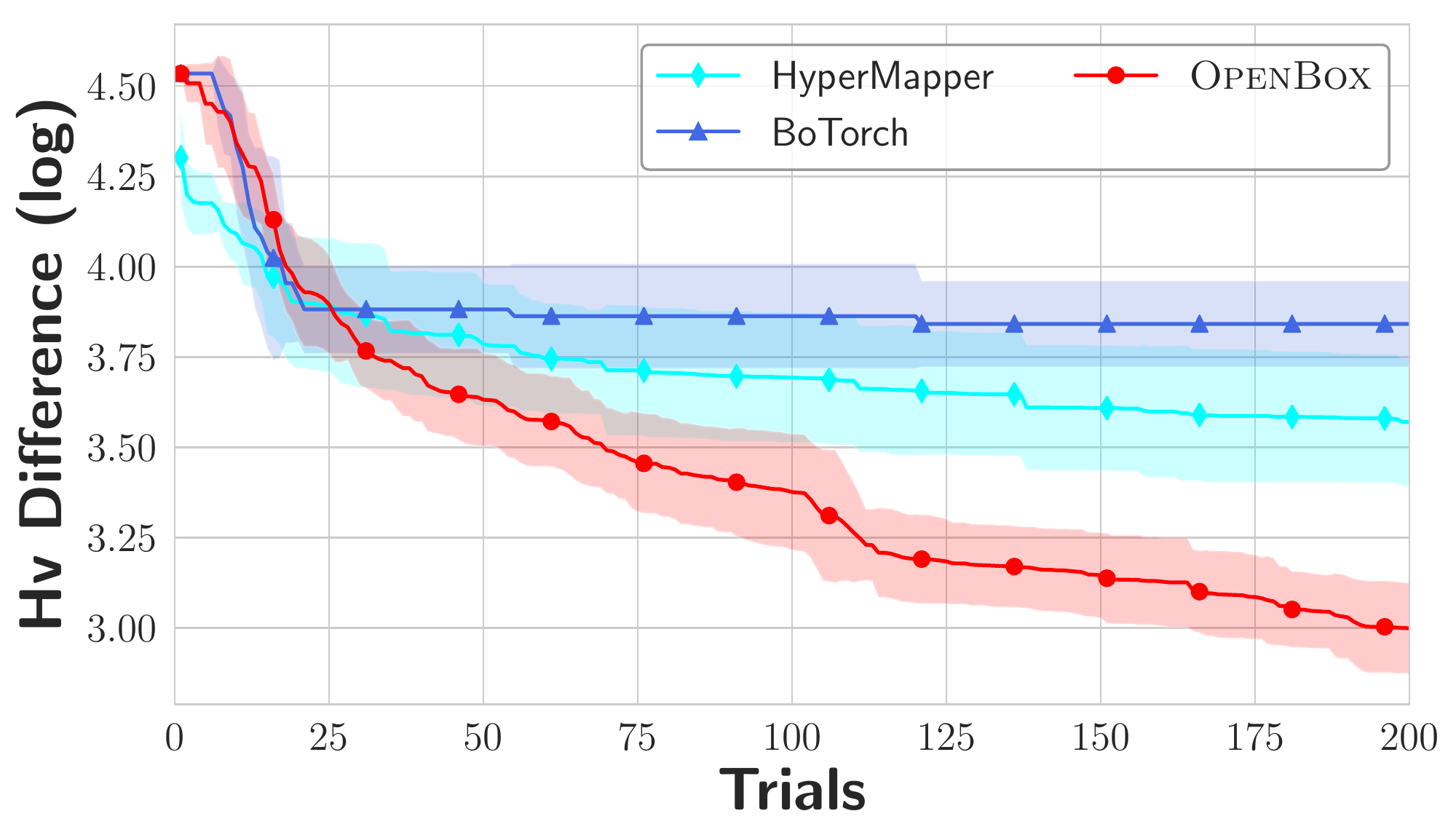}
			\label{moc_srn}
	}}
	\vspace{-1.5em}
	\caption{Results on multi-objective problems without (a and b) and with (c and d) constraints.}
	\vspace{-1.5em}
  \label{fig:mobo}
\end{figure*}

\subsubsection{Single-Objective Problems with Constraints}
Figure \ref{fig:soboc} shows the results of \sys along with the baselines on four constrained single-objective problems.
Besides \texttt{Random Search}, we compare \sys with three of the software packages that support constraints. 
\sys surpasses all the considered baselines on the convergence result. 
Note that on the 10-dimensional \texttt{Keane} problem in which the ground-truth optimal value is hard to locate, \sys is the only method that successfully optimizes this function while the other methods fail to suggest sufficient feasible configurations.

\subsubsection{Multi-Objective Problems without Constraints}
We compare \sys with three baselines that support multiple objectives and the results are depicted in Figure \ref{mo_zdt2} and \ref{mo_dtlz1}. 
In Figure \ref{mo_zdt2}, the hypervolume difference of \texttt{GPflowOpt} and \texttt{Hypermapper} decreases slowly as the number of trials grow, while \texttt{BoTorch} and \sys obtain a satisfactory Pareto Front quickly within 50 trials. 
In Figure \ref{mo_dtlz1} where the number of objectives is 5, \texttt{BoTorch} meets the bottleneck of optimizing the Pareto front while \sys tackles this problem easily by switching its inner algorithm from EHVI to MESMO; \texttt{GPflowOpt} is missing due to runtime errors.

\subsubsection{Multi-Objective Problems with Constraints}
We compare \sys with \texttt{Hypermapper} and \texttt{BoTorch} on constrained multi-objective problems (See Figure \ref{moc_constr} and \ref{moc_srn}). Figure \ref{moc_constr} demonstrates the performance on a simple problem, in which the convergence result of \sys is slightly better than the other two baselines. However, in Figure \ref{moc_srn} where the constraints are strict, \texttt{BoTorch} and \texttt{Hypermapper} fail to suggest sufficient feasible configurations to update the Pareto Front.
Compared with \texttt{BoTorch} and \texttt{Hypermapper}, \sys has more stable performance when solving multi-objective problems with constraints.

\subsection{Results on AutoML Tuning Tasks}
\subsubsection{AutoML Tuning on 25 OpenML datasets}
Figure \ref{fig:automl_rank} demonstrates the universality and stability of \sys in 25 AutoML tuning tasks. 
We compare \sys with \texttt{SMAC3} and \texttt{Hyperopt} on \texttt{LibSVM} since only these two baselines support \texttt{CATEGORICAL} parameters with conditions.
In general, \sys is capable of handling different types of input parameters while achieving the best median performance among the baselines considered.

\begin{figure}[t]
    \vspace{-0.6em}
	\centering
	\subfigure[AutoML Benchmark on LightGBM]{
		\scalebox{0.47}[0.47]{
			\includegraphics[width=1\linewidth]{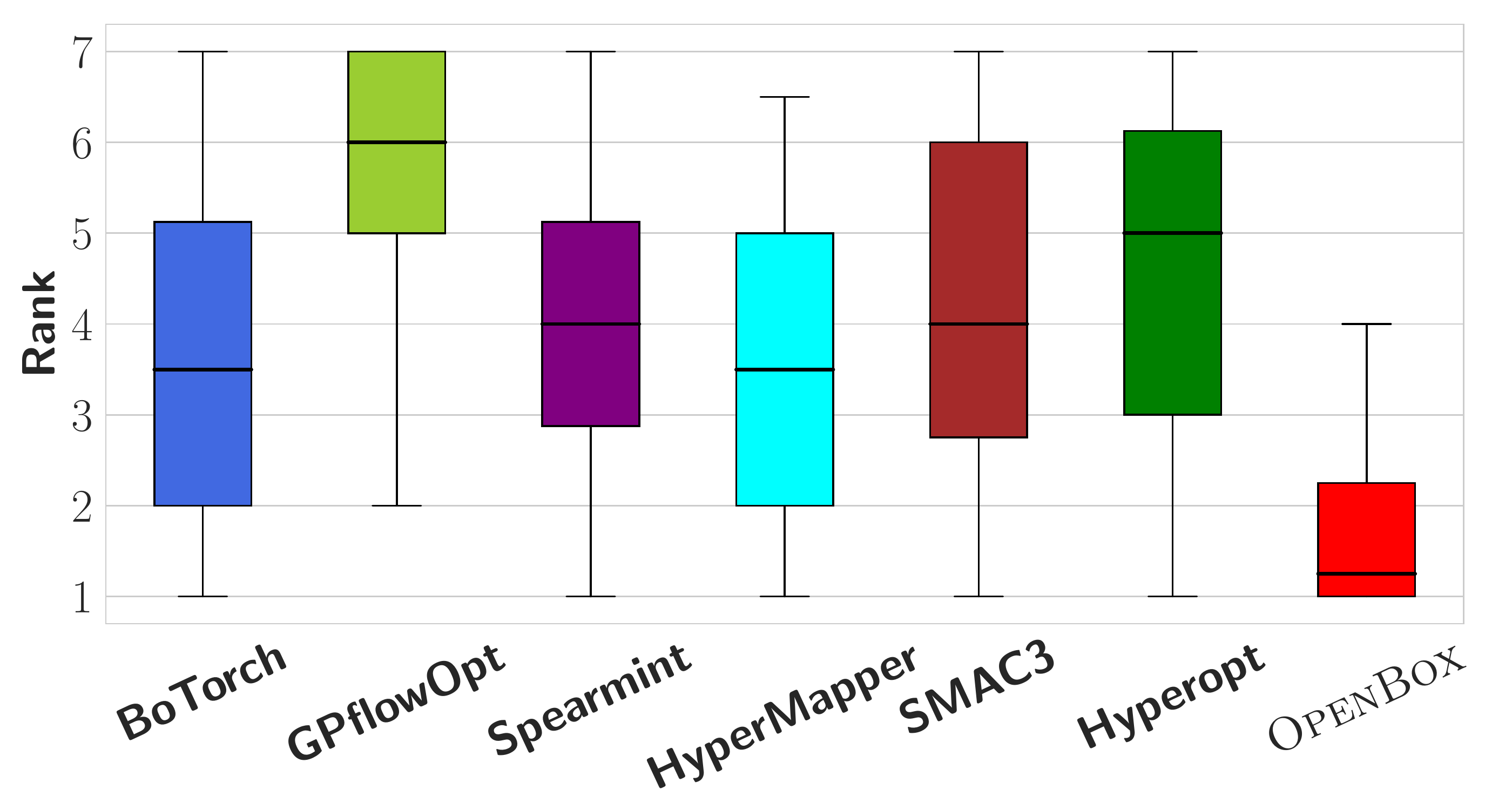}
	}}
		\subfigure[AutoML Benchmark on LibSVM]{
		\scalebox{0.47}[0.47]{			\includegraphics[width=1\linewidth]{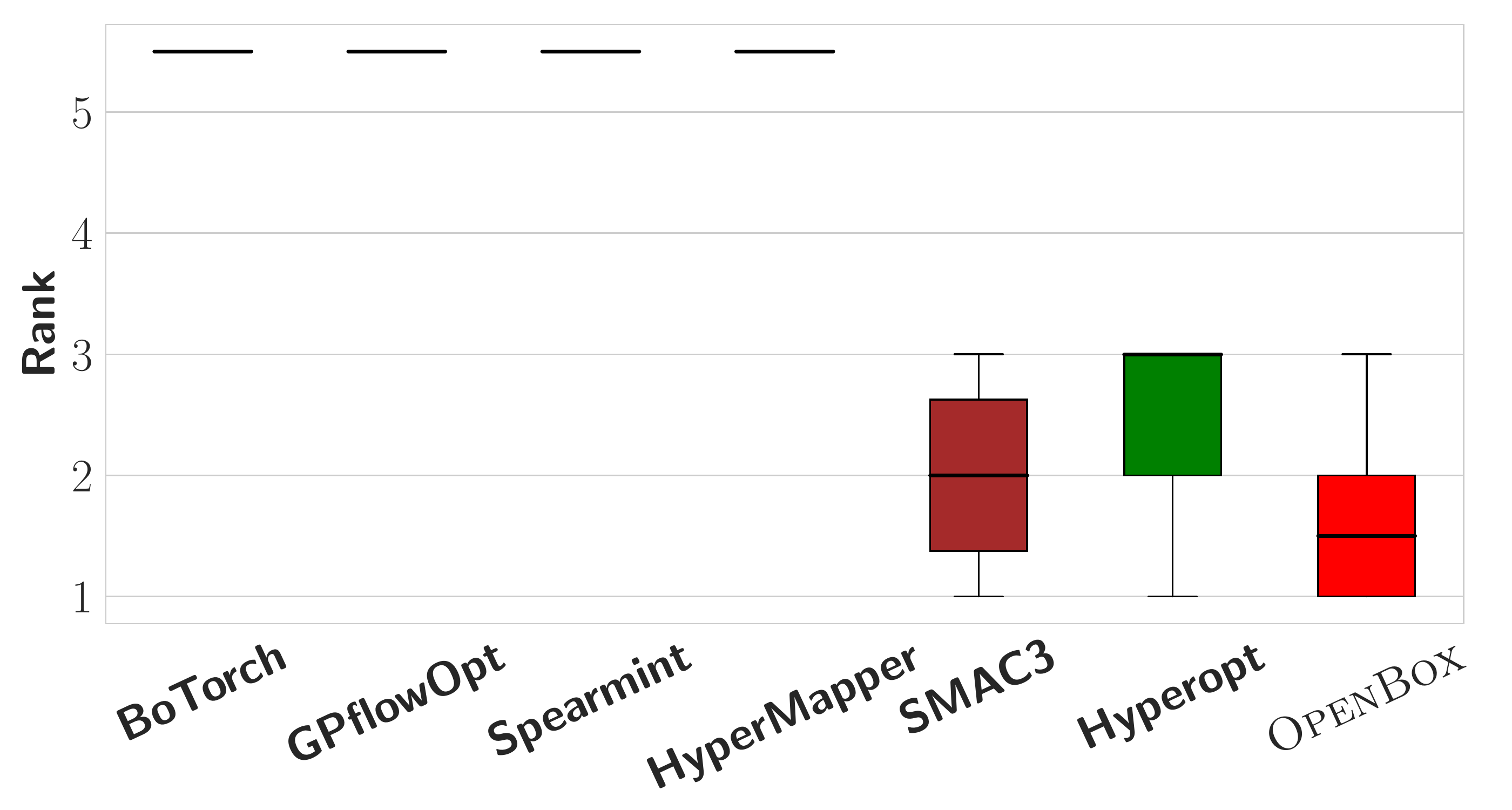}
	}}
	\vspace{-1.5em}
	\caption{Performance rank on 25 datasets (the lower is the better). The box extends from the lower to upper quartile values, with a line at the median. The whiskers extend from the box to show the range of the data.}
  \label{fig:automl_exp}
\end{figure}

\subsubsection{Parallel Experiments}
\begin{figure}[t]
    \vspace{-0.6em}
	\centering
	\subfigure[Parallel Experiments on Optdigits]{
		\scalebox{0.47}[0.47]{
			\includegraphics[width=1\linewidth]{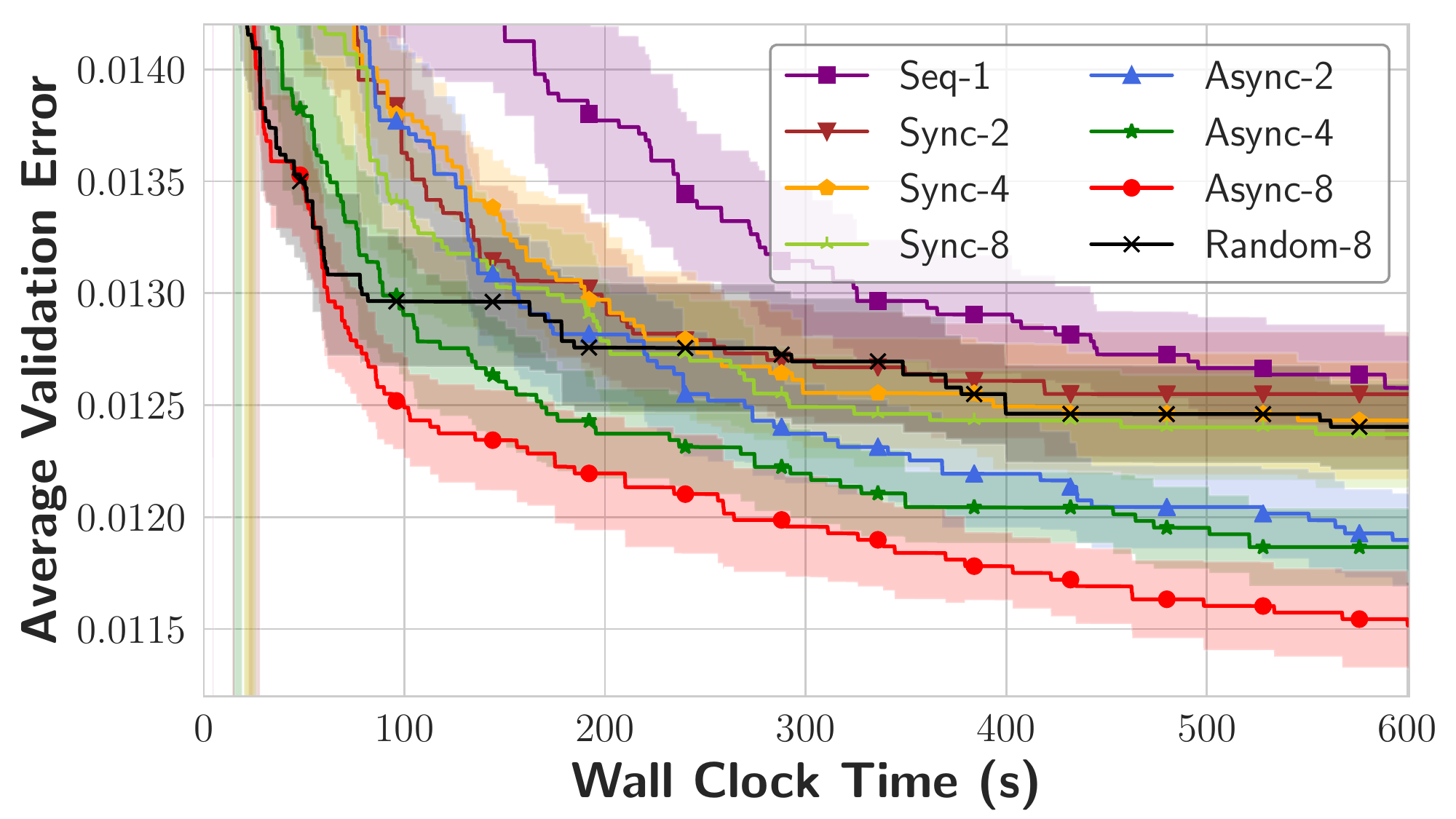}
		\label{fig:parallel_optidigits}
	}}
		\subfigure[Transfer Learning]{
		\scalebox{0.47}[0.47]{
			\includegraphics[width=1\linewidth]{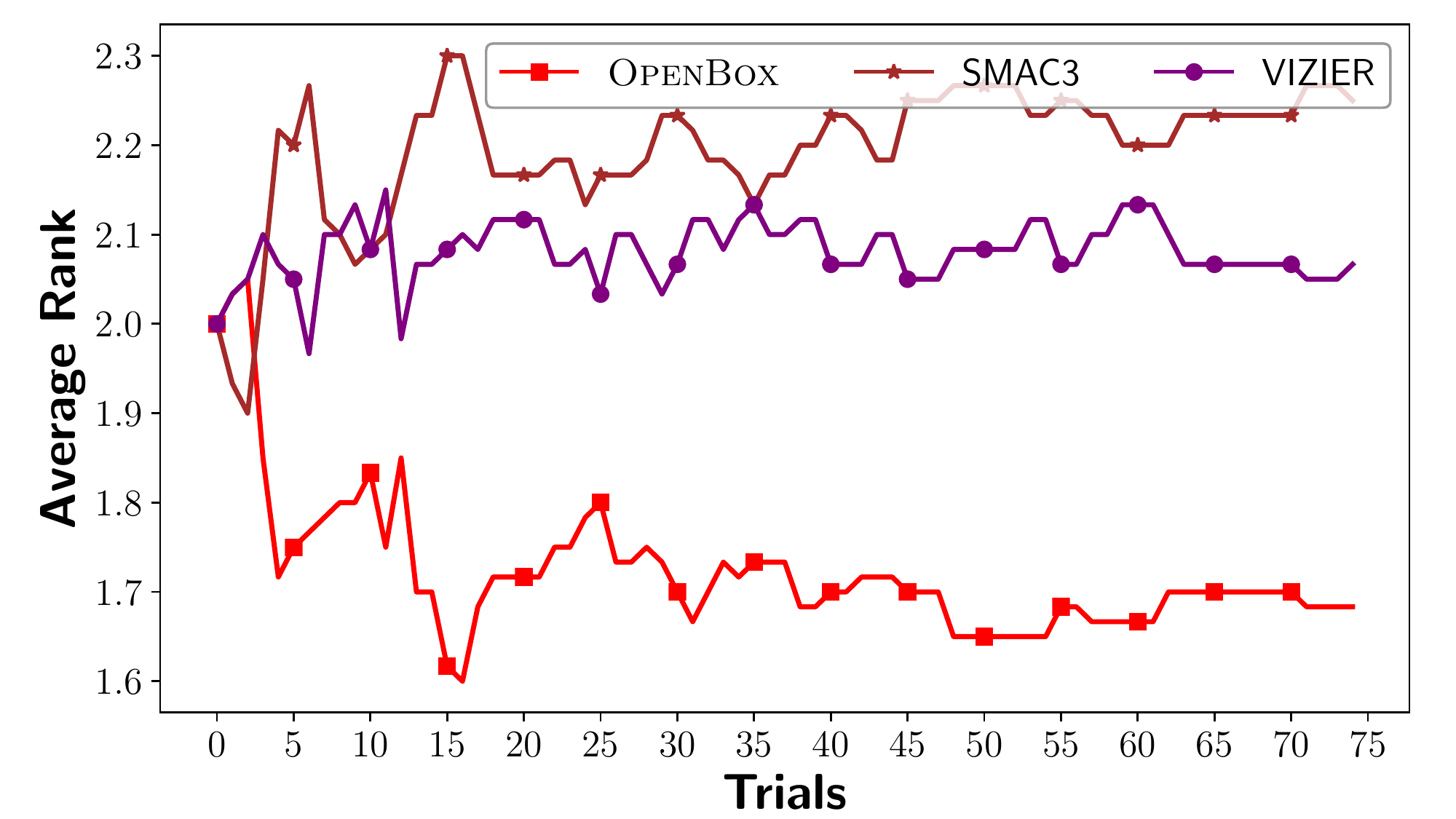}
		\label{fig:tl_exp}
	}}
	\vspace{-1.5em}
	\caption{Average validation error under two parallel settings (left figure) and average rank of tuning \texttt{LightGBM} with transfer learning (right figure). ``Seq'', ``Sync'' and ``Async'' refer to the sequential, sync and async mode respectively. The number of parallel workers is given after `-'. }
  \label{fig:automl_rank}
\end{figure}

To evaluate \sys with parallel settings, we conduct an experiment to tune the hyper-parameters of \texttt{LightGBM} on \texttt{Optdigits} with a budget of 600 seconds. 
Figure \ref{fig:parallel_optidigits} shows the average validation error with different parallel settings. 
We observe that the asynchronous mode with 8 workers achieves the best results and outperforms \texttt{Random Search} with 8 workers by a wide margin. 
It brings a speedup of $8\times$ over the sequential mode, which is close to the ideal speedup. 
In addition, although the synchronous mode brings a certain improvement over the sequential mode in the beginning, the convergence result is usually worse than the asynchronous mode due to stragglers.

\subsubsection{Transfer Learning Experiment}
In this experiment, we remove all baselines except \texttt{Vizier}, which provides the transfer learning functionality for the traditional black-box optimization. 
We also add \texttt{SMAC3} that provides a non-transfer reference. 
In addition, this experiment involves tuning \texttt{LightGBM} on 25 OpenML datasets, and it is performed in a leave-one-out fashion, i.e, we tune the hyperparameters of \texttt{LightGBM} on a dataset (target problem), while taking the tuning history on the remaining datasets as prior observations.
Figure~\ref{fig:tl_exp} shows the average rank for each baseline. 
We observe that 1) \texttt{Vizier} and \sys show improved sample efficiency relative to \texttt{SMAC3} that cannot use prior knowledge from source problems, and 2) the proposed transfer learning framework in \sys performs better than the transfer learning algorithm used in \texttt{Vizier}.
Furthermore, it is worth mentioning that \sys also supports transfer learning for the generalized black-box optimization, while \texttt{Vizier} does not.

\section{Conclusion}

In this paper, we have introduced a service that aims for solving generalized BBO problems -- \sys, which is open-sourced and highly efficient. We have presented new principles from a service perspective that drive the system design, and we have proposed efficient frameworks for accelerating BBO tasks by leveraging local-penalization based parallelization and transfer learning.
\sys hosts lots of state-of-the-art optimization algorithms with consistent performance, via adaptive algorithm selection.
It also offers a set of advanced features, such as performance-resource extrapolation, multi-fidelity optimization, automatic early stopping, and data privacy protection.
Our experimental evaluations have also showcased the performance and efficiency of \sys on a wide range of BBO tasks.

\begin{acks}
This work is supported by the National Key Research and Development Program of China (No.2018YFB1004403), NSFC (No.61832001, U1936104), Beijing Academy of Artificial Intelligence (BAAI), and Kuaishou-PKU joint program. Bin Cui is the corresponding author. 
\end{acks}

\bibliographystyle{ACM-Reference-Format}
\bibliography{reference}


\clearpage
\appendix

\section{Appendix}

\subsection{Performance-Resource Extrapolation}
While optimizing various black-box problems, we observe that the optimization curve (performance vs. trials) is often \textit{saturating}, i.e., after a certain number of trials, more evaluations will not cause a meaningful improvement $\delta > 0$ in performance. \sys applies a combined learning curve extrapolation method inspired by \cite{Domhan2015}, which early stops the training procedure of neural networks when the performance of the network becomes less likely to improve.

We measure the performance by negative hypervolume indicator (HV) of the Pareto set $\mathcal{P}$ bounded above by reference point $r$, denoted by $HV(\mathcal{P}, r)$. In single-objective case, $\mathcal{P} = \{y_{\text{best}}\}$. Note that in both cases, the performance is \textit{decreasing}.

Denote the performance at timestep $t$ by $z_t$. Given observed data $z_{1:n} := \{z_1, \ldots, z_n\}$, a natural idea is to estimate whether the performance at a future timestep $t > n$ will exceed the current best performance $z_n$. 
We extrapolate the performance curve $z_t$ with a weighted probabilistic model 
\[
g_{\text{comb}}(t|\bm{\Theta}) = \sum_{k=1}^K w_k g_k(t|\bm{\theta_k}) + \varepsilon,
\]
where each of $g_1, \ldots, g_K$ is a parametric family of decreasing saturating functions, and $\varepsilon \sim \mathcal{N}(0, \sigma^2)$. We estimate $\bm{\Theta} = (w_1, \ldots, w_K,$ $\theta_1, \ldots, \theta_K, \sigma^2)$ using Markov Chain Monte Carlo (MCMC) inference. The prior and posterior distribution over $\bm{\Theta}$ are as follows
\[
p(\bm{\Theta}) \propto \Big( \prod_{k=1}^K p(w_k) p(\bm{\theta_k}) \Big) p(\sigma^2) \mathbbm{1}(g_{\text{comb}}(1 | \bm{\Theta}) > g_{\text{comb}}(t | \bm{\Theta})),
\]
\[
P(\bm{\Theta} | z_{1:n}) \propto P(z_{1:n} | \bm{\Theta}) P(\bm{\Theta}),
\]
where $t > n$.

We sample $\bm{\Theta}$ from the posterior and compute $P(z_t < z_n - \delta |z_{1:n})$, which is the probability that the optimization procedure yields a meaningful improvement $\delta$ at timestep $t$.

\subsection{Bayesian Optimization Algorithms}
The BO algorithms in \sys include three parts: surrogate models, acquisition functions, and acquisition function optimizers. 
Note that, partial implementations for single-objective BBO without constraints, including probabilistic random forest surrogate, ei optimization, are inheriting from the SMAC3\footnote{\url{https://github.com/automl/SMAC3}} package directly.

\vspace{-0.5em}
\paragraph*{Surrogate Models}
\sys selects different surrogate models based on the number of trials.
For tasks with under 500 trials, \sys defaults to using Gaussian Process (GP) from \texttt{scikit-optimize} package. 
We use a Mat\'ern kernel with automatic relevance determination (ARD) for continuous parameters and a Hamming kernel for categorical parameters. 
When both continuous and categorical parameters exist, we use the product of these two kernels. 
The parameters of GP are fitted by optimizing the marginal log-likelihood with the gradient-based method (as default) or MCMC sampling.
Due to the high computational complexity $\mathcal{O}(n^3)$, GP cannot scale well to the setting with too many trials (a large $n$). 
Therefore, for tasks with more than 500 trials, the surrogate model is switched to probabilistic random forest proposed in \cite{hutter2011sequential}, which incurs less complexity. 

\vspace{-0.5em}
\paragraph*{Acquisition Functions}
By default, \sys uses Expected Improvement (EI)~\cite{movckus1975bayesian} for single-objective optimization, Expected Hypervolume Improvement (EHVI)~\cite{Emmerich2005} for multi-objective optimization, and Probability of Feasibility (PoF)~\cite{Gardner2014} for constraints.
\sys computes these acquisition functions analytically~\cite{yang2019multi} (by default) or through Monte Carlo integration~\cite{daulton2020differentiable}. 
In addition, \sys includes multiple acquisition functions to meet the needs of different problem settings. For single-objective optimization, Expected Improvement per second (EIPS)~\cite{snoek2012practical} can be used to find a good configuration as quickly as possible, and Expected Improvement with Local Penalization (LP-EI)~\cite{Gonzalez2016} utilizes local penalizers to propose batches of configurations simultaneously. For multi-objective optimization, Max-value Entropy Search for Multi-objective Optimization (MESMO)~\cite{Belakaria2019} and Uncertainty-aware Search framework~\cite{belakaria2020uncertainty} for Multi-objective Optimization (USeMO) work efficiently when the number of objectives is large. Other implemented acquisition functions include Probability of Improvement (PI), and Upper Confidence Bound (UCB)~\cite{Srinivas2010}.  

\vspace{-0.5em}
\paragraph*{Acquisition Function Optimizers}
To support generic surrogate models that are not differentiable, we maximize the acquisition function via the following two methods: 1) interleaved local and random search (gradient-free) which can handle categorical parameters, and 2) multi-start staged optimizer of random search and L-BFGS-B from \texttt{Scipy} (estimate gradient by 2-point finite difference) which can locate the global optimum in high dimensional design space efficiently.

\subsection{Transfer Learning Details}
In \sys, we expand RGPE~\cite{feurer2018scalable}, a state-of-the-art transfer learning method on single-objective problems, into generalized settings. 

First, for each prior task $i$, we train surrogates $M^i_{1:m}$ for $m$ objectives on the corresponding observations from $D^i$. 
Then we build surrogates $M^{\text{TL}}_{1:m}$ to guide the optimization instead of using the original surrogates $M^T_{1:m}$ fitted on $D^{T}$ only. 
For ease of description, we assume there is only one surrogate $M^{\text{TL}}$ since the method of building surrogate for each objective is exactly the same.
The prediction of $M^{\text{TL}}$ at point $\bm{x}$ is given by $y \sim \mathcal{N}(\sum_{i}\mu_{\text{TL}}(\bm{x}), \sigma^2_{\text{TL}}(\bm{x}))$, where

\begin{equation}
\begin{aligned}
\mu_{\text{TL}}(\bm{x})&=(\sum_i\mu_i(\bm{x}) {\bf w}_i \sigma_i^{-2}(\bm{x}))\sigma_{\text{TL}}^2(\bm{x}),\\
\sigma^2_{\text{TL}}(\bm{x})&=(\sum_i {\bf w}_i \sigma_i^{-2}(\bm{x}))^{-1},\\
\end{aligned}
    \nonumber
\end{equation}
where ${\bf w}_i$ is the weight of base surrogate $M^i$, and $\mu_i$ and $\sigma^2_i$ are the predictive mean and variance from base surrogate $M^i$.
The weight ${\bf w}_i$ reflects the similarity between the previous task and current task.
Therefore, $M^{\text{TL}}$ carries the knowledge of the prior tasks, which could greatly accelerate the convergence of the optimization on the current task.
We then use the following ranking loss function $L$, i.e., the number of misranked pairs, to measure the similarity between previous tasks and current task:
\begin{equation}
    L(M^i,D^T) = \sum_{j=1}^{n^{T}}\sum_{k=1}^{n^{T}}\mathbbm{1}((M^i(\bm{x}_j) < (\bm{x}_k) \oplus (y_j < y_k)),
    \label{rank_loss}
\end{equation}
where $\oplus$ is the exclusive-or operator, $n^T=|D^T|$, $\bm{x}_j$ and $y_j$ are the sampled point and its performance in $D^T$, and $M^i(\bm{x}_j)$ means the prediction of $M^i$ on the point $\bm{x}_j$.
Based on the ranking loss function, the weight ${\bf w}_i$ is set to the probability that $M^i$ has the smallest ranking loss on $D^T$, that is,
${\bf w}_i = P(i=\operatorname{argmin}_{j}L(M^j, D^T))$. This probability can be estimated using the MCMC sampling.

\subsection{Discussions about Local Penalization based Parallelization}
Algorithm~\ref{algo:paralllel_sample} parallelizes BO algorithms by imputing the configurations being evaluated with the median of the evaluated data $D_n = \{\bm{x_i}, \bm{y_i}\}_{i=1}^n$. 
For notational simplicity, we discuss the single-objective case with EI as acquisition function. Denote the median of observed values $\{y_i\}_{i=1}^n$ by $\hat{y}$, and the smallest observed value by $\eta$. Define $u = f(\bm{x}),\ u \sim \mathcal{N}(\mu_n(\bm{x}), \sigma_n^2(\bm{x}))$, where $\mu_n(\bm{x})$ and $\sigma_n^2(\bm{x})$ are the mean and variance of the posterior distribution of the surrogate model trained on $D_n$.
The expected improvement is
\begin{align}
\begin{split}
\alpha_{\text{EI}}(\bm{x}; D_n) &= \mathbb{E}_u [(\eta - u) \mathbbm{1}(u < \eta)] \\
&= (\eta - \mu_n(\bm{x})) \Phi(z) + \sigma_n(\bm{x}) \phi(z)
\end{split}
\end{align}
when $\sigma_n > 0$ and vanishes otherwise. Here, $\Phi$ and $\phi$ are the CDF and PDF of the standard normal distribution, $z = \frac{\eta - \mu_n(\bm{x})}{\sigma_n(\bm{x})}$.

We first show that, with our imputation strategy,  $\alpha_{\text{EI}}(\bm{x}; D_{\text{aug}})$ will be sufficiently small if $\bm{x}$ is close to some $\bm{x}_{\text{eval}} \in D_{\text{aug}}$, i.e., locally penalized near $\bm{x}_{\text{eval}}$. 
For all probabilistic surrogate models, $\mu_n(\bm{x}) = f(\bm{x}),\ \sigma_n(\bm{x}) = 0$ if $\bm{x} \in D_n$, which means $\alpha_{\text{EI}}(\bm{x}) = 0,\ \forall \bm{x} \in D_n$. By augmenting $D_n$ with $D_{\text{new}} = \{(\bm{x}_{\text{eval}}, \hat{y}): \bm{x}_{\text{eval}} \in C_{\text{eval}}\}$, we have $\alpha_{\text{EI}}(\bm{x}_{\text{eval}}) = 0,\ \forall \bm{x}_{\text{eval}} \in C_{\text{eval}}$.  Since $\alpha_{EI}(\bm{x}; D_{\text{aug}})$ is continuous if the surrogate is GP and flat if the surrogate is random forest, when $\bm{x}$ is close to some $\bm{x}_{\text{eval}} \in C_{\text{eval}}$, $\eta - \mu_n(\bm{x}) \approx \eta - \hat{y}$ and $z = (\eta - \mu_n(\bm{x}))/ \sigma_n(\bm{x}) $ are negative and sufficiently small. Hence, both terms in (2) are small and $\bm{x}$ is unlikely to be the maximum of $\alpha_{\text{EI}}$. This conclusion can be naturally extended to cases with multiple objectives, and more generally, other acquisition functions.

Moreover, although Algorithm~\ref{algo:paralllel_sample} changes the posterior distribution of the surrogate by imposing a local penalty, it helps avoid over-exploitation. Considering the configurations evaluated at the same time as a "batch", Algorithm~\ref{algo:paralllel_sample} simplified the complex joint optimization problem by assigning a different region for each worker to explore. From the experiment results shown in Figure \ref{fig:parallel_optidigits}, we observe that Algorithm~\ref{algo:paralllel_sample} is a highly efficient, as well as widely applicable parallelization heuristic.

\subsection{More Experimental Results}
\begin{figure}[htb]
	\centering
	\subfigure[LightGBM on Puma32H]{
		\scalebox{0.46}{
			\includegraphics[width=1\linewidth]{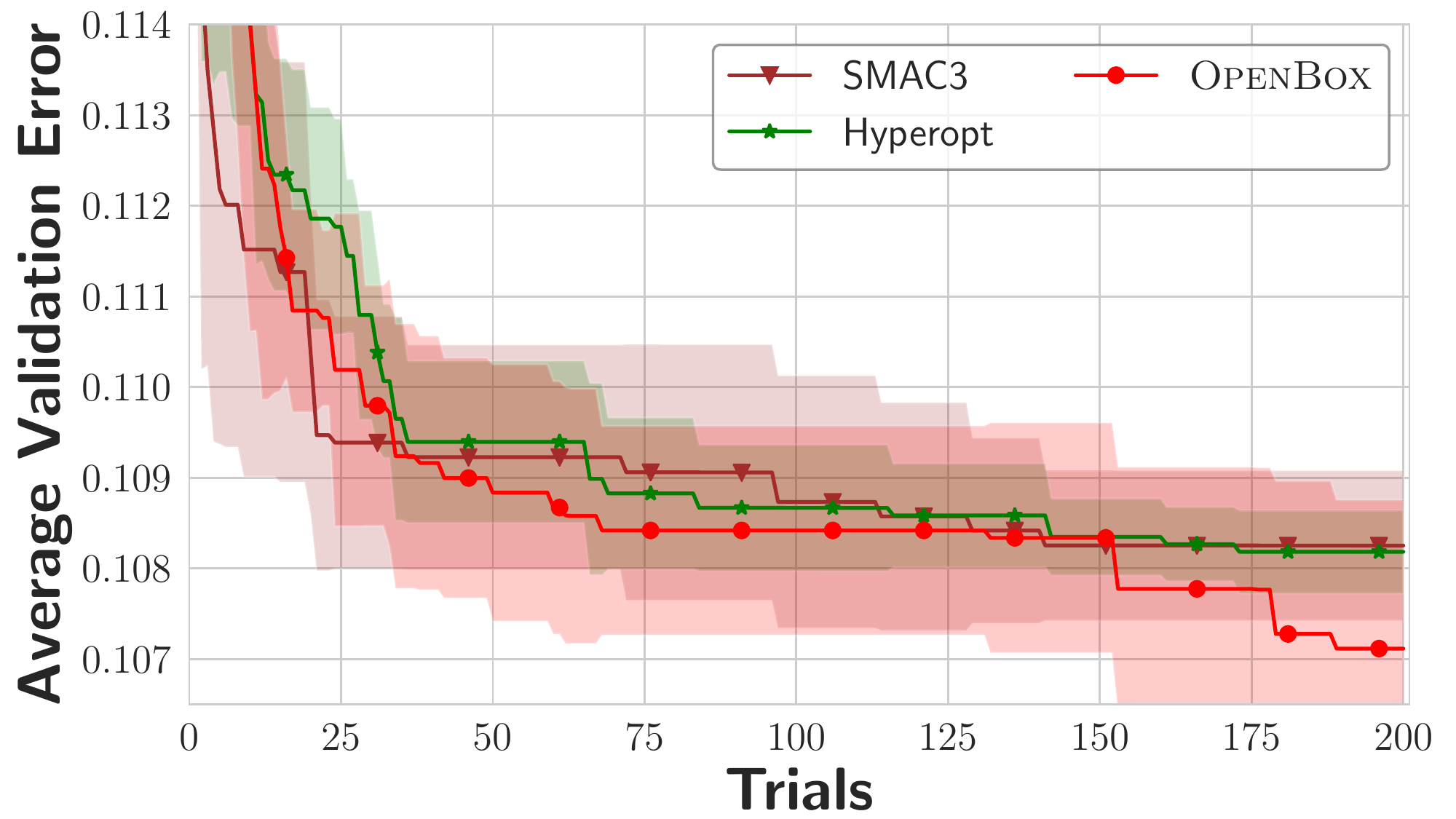}
			\label{puma32H}
	}}
	\subfigure[LightGBM on Puma8NH]{
		\scalebox{0.46}{
			\includegraphics[width=1\linewidth]{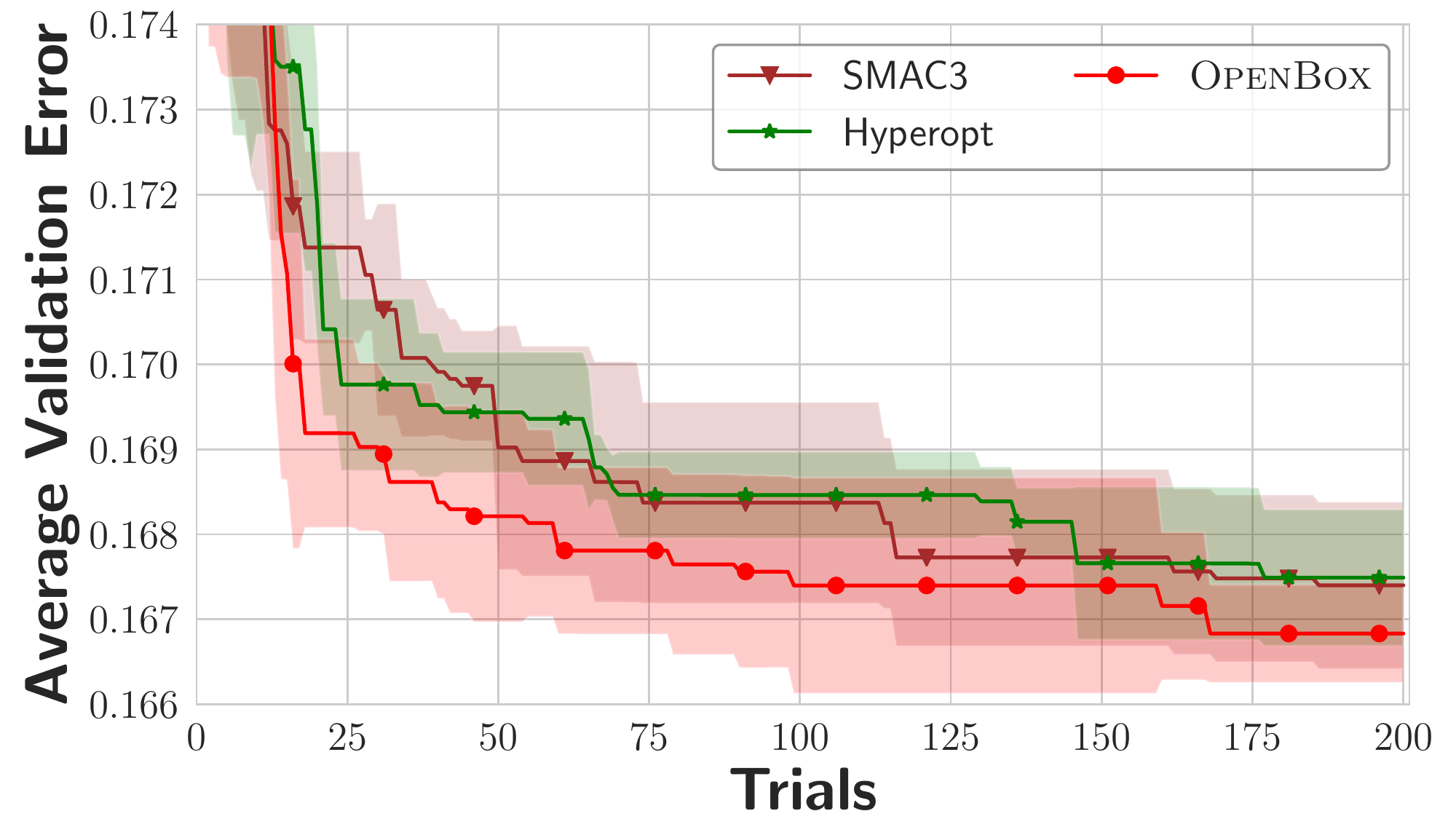}
			\label{puma8NH}
	}}
	\subfigure[LibSVM on Pollen]{
		\scalebox{0.46}{
			\includegraphics[width=1\linewidth]{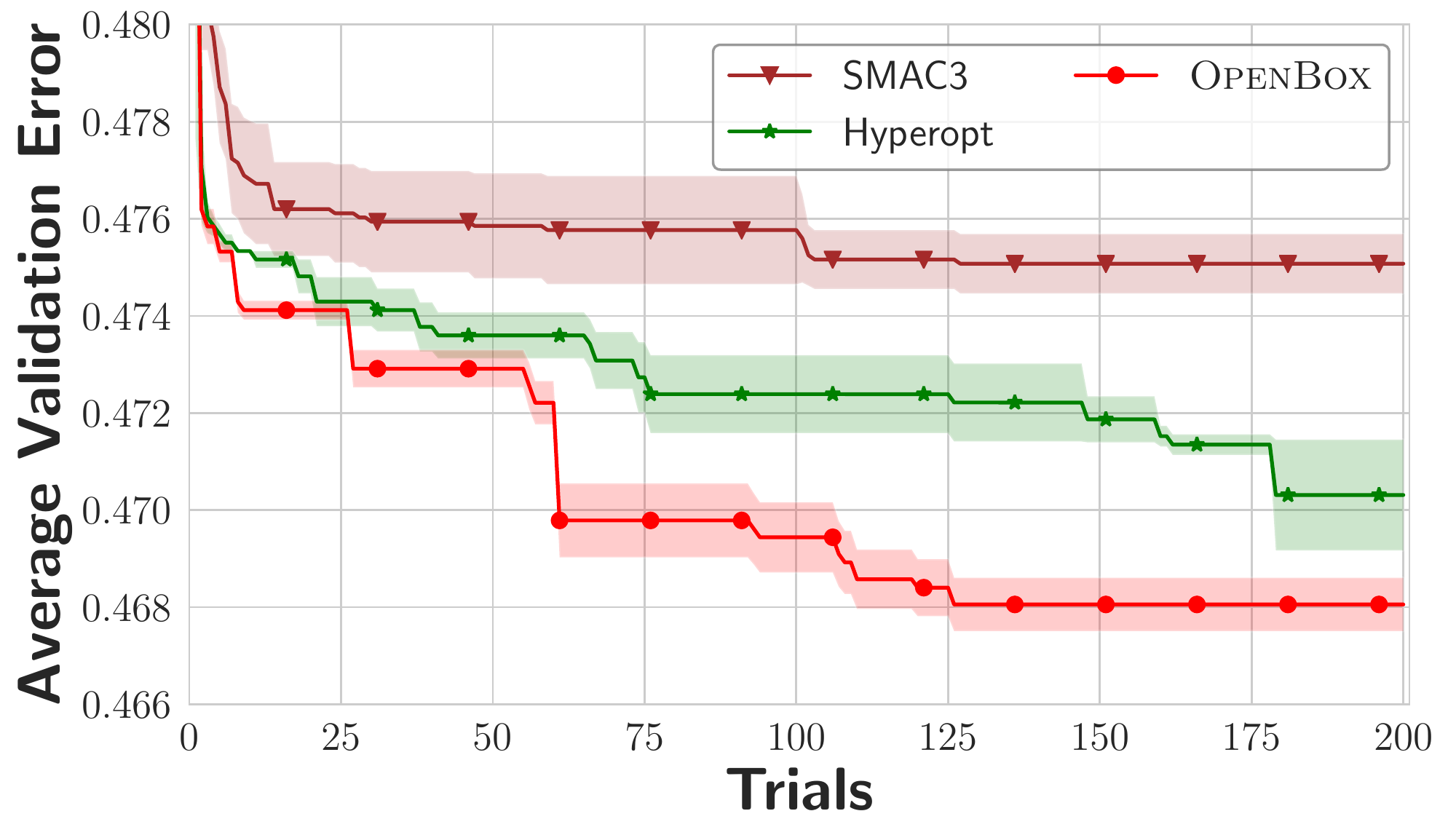}
			\label{pollen}
	}}
	\subfigure[LibSVM on Wind]{
		\scalebox{0.46}{
			\includegraphics[width=1\linewidth]{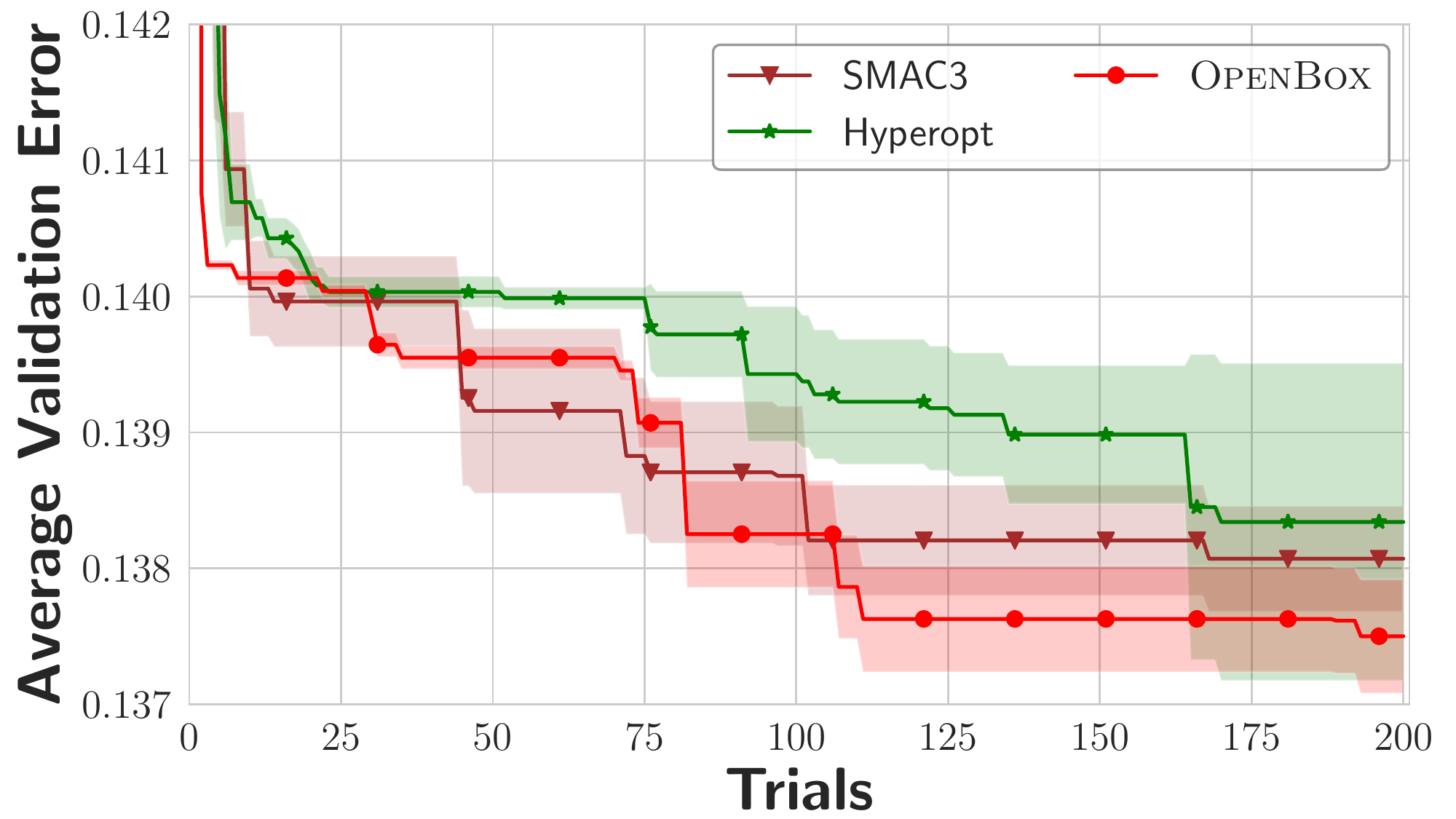}
			\label{wind}
	}}
	\vspace{-1.5em}
	\caption{Performance of two AutoML tasks on 4 datasets.}
  \label{fig:automl performance}
\end{figure}

\paragraph{AutoML Performance.} 
Besides the rank of convergence results shown in Figure \ref{fig:automl_rank}, we present Figure \ref{fig:automl performance} that demonstrates the optimization process of \sys on AutoML tasks. \sys achieves 2.0-3.3$\times$ speedups over the best baseline in each task.

\begin{figure}[htb]
\centering
\vspace{-1em}
\scalebox{0.6}[0.6]{
\includegraphics[width=0.5\textwidth]{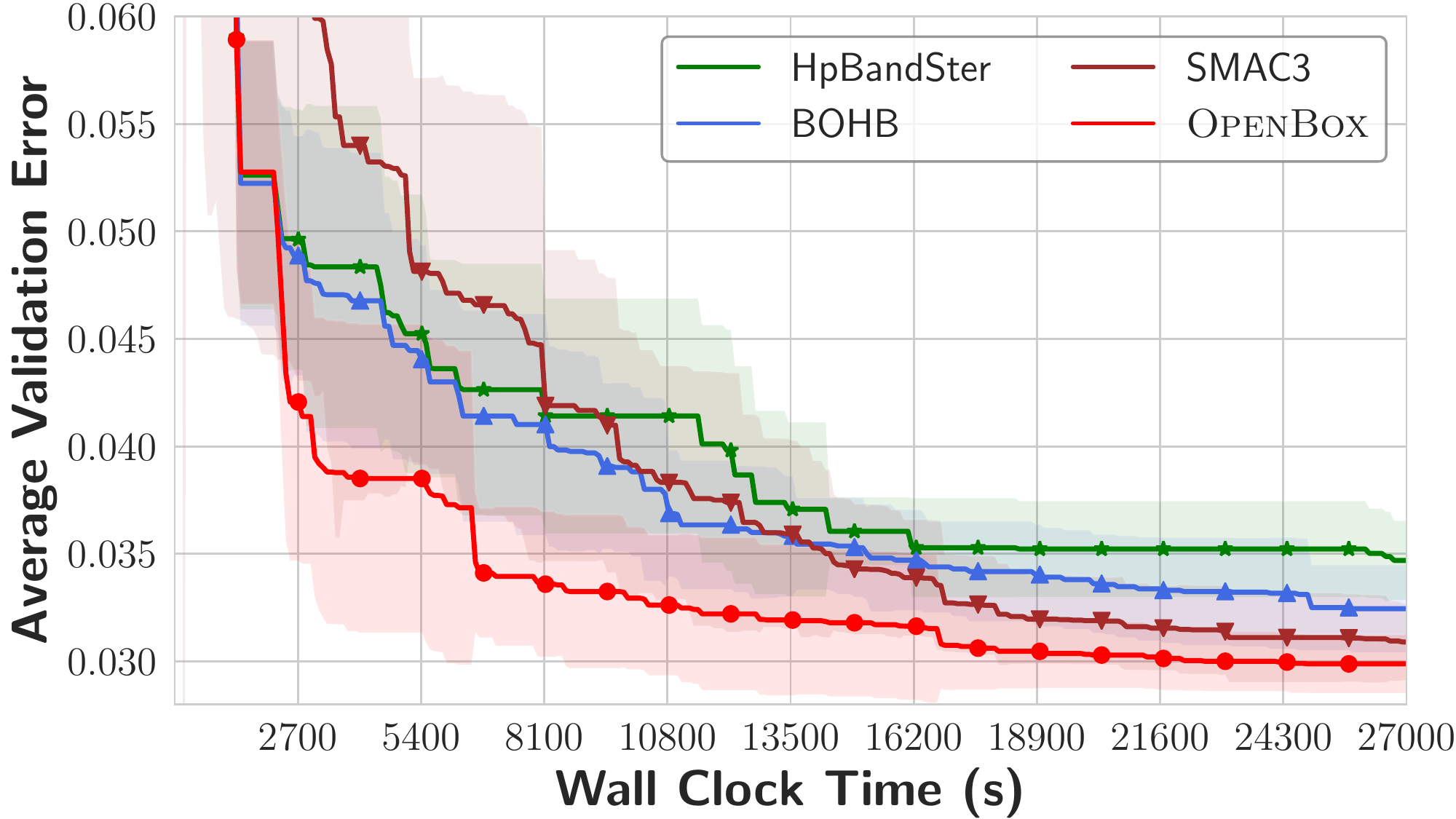}
}
\vspace{-1em}
\caption{Multi-fidelity experiment on tuning hyper-parameters of LightGBM.}
\label{fig:mfbo}
\end{figure}

\paragraph{Muiti-fidelity Acceleration.}
Figure \ref{fig:mfbo} shows the acceleration of \sys using multi-fidelity optimization compared with \texttt{SMAC3} and two other multi-fidelity packages, \texttt{HpBandSter} and \texttt{BOHB}. The dataset used in this experiment is \texttt{Covtype}, which is a large-scale dataset with over 580k samples. We observe that though \texttt{HpBandSter} and \texttt{BOHB} accelerates the optimization in the beginning, their convergence results are worse than that of \texttt{SMAC3}. However, \sys obtains a 3.8 $\times$ speedup over \texttt{SMAC3} when achieving the comparable convergence performance.

\subsection{Reproduction Instructions}
We run our experiments on 2 machines with 56 Intel(R) Xeon(R) CPU E5-2680 v4 @ 2.40GHz. The versions of baselines are 
1) \texttt{BoTorch} 0.3.3, 
2) \texttt{GPflowOpt} 0.1.1, 
3) \texttt{HyperMapper} master branch~\footnote{https://github.com/luinardi/hypermapper},
4) \texttt{SMAC3} 0.8.0, 
5) \texttt{Hyperopt} 0.2.3 and 6) \texttt{Spearmint} master branch~\footnote{https://github.com/JasperSnoek/spearmint}. 
The source code of \sys is written in Python 3.7 and is already available in Github~\footnote{https://github.com/PKU-DAIR/open-box}. We place the code for reproduction under the directory \textit{test/reproduction}. 
For example, to run single-objective experiment on Branin, the script is as follows:

\vspace{0.5em}
\noindent
\texttt{python test/reproduction/so/benchmark\_xxx.py \\ --problem branin --n 200}.

\end{document}